\RequirePackage{fix-cm}
\documentclass[twocolumn]{svjour3}    
\pdfoutput=1

\usepackage{graphicx}
\usepackage{natbib}
\usepackage{url}            
\usepackage{booktabs}       
\usepackage{amsfonts}       
\usepackage{nicefrac}       
\usepackage{microtype}      
\usepackage{graphicx}
\usepackage{enumitem}
\usepackage{bm}
\usepackage{nccmath}
\usepackage{boldline}
\usepackage{booktabs}        
\usepackage{threeparttable}

\usepackage[noend]{algcompatible}

\usepackage{capt-of}
\usepackage{color,soul}
\usepackage[pagebackref=false,breaklinks=true,colorlinks,bookmarks=false,urlcolor=magenta, citecolor=blue]{hyperref}

\usepackage{times}
\usepackage{epsfig}
\usepackage{multirow}
\usepackage{amsmath}
\usepackage{amssymb}
\usepackage{bbding}
\usepackage{wrapfig}
\usepackage{comment}
\usepackage{pifont}
\usepackage[table]{xcolor}
\usepackage{bm}
\usepackage{soul}

\usepackage{stfloats}
\usepackage{cuted}
\usepackage{capt-of}
\usepackage{graphicx}
\usepackage{booktabs}
\usepackage{enumitem}
\usepackage{amsfonts}
\usepackage{bbm}
\usepackage{textcomp}

\usepackage[ruled,linesnumbered]{algorithm2e}
\usepackage{makecell}
\usepackage{subfigure}
\usepackage{arydshln}
\usepackage{ulem}

\def\cm#1{\checkmark}

\useunder{\underline}{\ul}{}

\usepackage{tikz}

\begin{document}
\begin{sloppypar}
\title{AgMTR: Agent Mining Transformer for Few-shot Segmentation in Remote Sensing}

\author{Hanbo Bi$^{1,2,3,4}$ \and 
Yingchao Feng$^{1,4\star}$ \and
Yongqiang Mao$^{1,2,3,4,5}$ \and \\
Jianning Pei$^{1,2,3,4}$ \and 
Wenhui Diao $^{1,4}$ \and
Hongqi Wang $^{1,4}$ \and
Xian Sun $^{1,2,3,4}$ 
}

\institute{
Hanbo Bi (bihanbo21@mails.ucas.edu.cn) \\ 
Yingchao Feng (fengyc@aircas.ac.cn) \\
Yongqiang Mao (mao.yongqiang@ieee.org) \\
Jianning Pei (peijiangning22@mails.ucas.ac.cn) \\
Wenhui Diao (diaowh@aircas.ac.cn)\\
Hongqi Wang (wiecas@sina.com)\\ 
Xian Sun (sunxian@aircas.ac.cn) \\
1. Aerospace Information Research Institute, Chinese Academy of Sciences \\
2. School of Electronic, Electrical and Communication Engineering, University of Chinese Academy of Sciences  \\
3. University of Chinese Academy of Sciences \\
4. Key Laboratory of Network Information System Technology, Institute of Electronics, Chinese Academy of Sciences \\
5. Department of Electronic Engineering, Tsinghua University\\
$\star$ Corresponding authors.\\
}


\date{Received: date / Accepted: date}

\maketitle

\begin{abstract}

Few-shot Segmentation (FSS) aims to segment the interested objects in the query image with just a handful of labeled samples (i.e., support images). Previous schemes would leverage the similarity between support-query pixel pairs to construct the pixel-level semantic correlation. However, in remote sensing scenarios with extreme intra-class variations and cluttered backgrounds, such pixel-level correlations may produce tremendous mismatches, resulting in semantic ambiguity between the query foreground (FG) and background (BG) pixels. To tackle this problem, we propose a novel Agent Mining Transformer (AgMTR), which adaptively mines a set of local-aware agents to construct agent-level semantic correlation. Compared with pixel-level semantics, the given agents are equipped with local-contextual information and possess a broader receptive field. At this point, different query pixels can selectively aggregate the fine-grained local semantics of different agents, thereby enhancing the semantic clarity between query FG and BG pixels. Concretely, the Agent Learning Encoder (ALE) is first proposed to erect the optimal transport plan that arranges different agents to aggregate support semantics under different local regions. Then, for further optimizing the agents, the Agent Aggregation Decoder (AAD) and the Semantic Alignment Decoder (SAD) are constructed to break through the limited support set for mining valuable class-specific semantics from unlabeled data sources and the query image itself, respectively. Extensive experiments on the remote sensing benchmark iSAID indicate that the proposed method achieves state-of-the-art performance. Surprisingly, our method remains quite competitive when extended to more common natural scenarios, i.e., PASCAL-$5^i$ and COCO-$20^{i}$.

\end{abstract}

\section{Introduction}
\label{Indroduction}
Semantic segmentation~\citep{dey2010review,diakogiannis2020resunet,mao2022beyond,yuan2021review,feng2021double}, one of the fundamental tasks in the intelligent interpretation of remote sensing, aims at assigning a pre-defined target category to each pixel in remote sensing images~\citep{wang2022empirical,wang2022self,mao2023elevation,zhang2022artificial,mao2024sdl}. With the recent development of deep learning techniques, especially the emergence of the fully convolutional network (FCN)~\citep{long2015fully,ronneberger2015u,he2016identity,chen2017deeplab}, the semantic segmentation task of remote sensing has made remarkable breakthroughs. However, collecting massive pixel-level annotations for complex remote sensing scenes is time-consuming and labor-intensive~\citep{csillik2017fast,kotaridis2021remote}. Even though researchers propose weakly-supervised~\citep{wang2020weakly,schmitt2020weakly,lian2021weakly} and semi-supervised methods~\citep{wang2020semi,sun2020bas,wang2022semi,zhang2022semi} to alleviate the data hunger, they fail to effectively generalize to unseen domains. Few-shot Learning (FSL) provides a promising research solution for tackling the above issue, which can exploit several labeled samples to quickly generalize to unseen (novel) classes~\citep{vinyals2016matching,bi2023not,sung2018learning,mao2022bidirectional,wang2020generalizing}.

Few-shot Segmentation (FSS) is a natural extension of FSL for handling dense prediction tasks~\citep{shaban2017one,zhang2019canet,xu2024taformer,xu2024taformer}. The purpose of FSS is to segment the corresponding object in the interested image (called query image) utilizing several labeled samples (called support images) containing the interested class. The current mainstream FSS schemes, i.e., prototype-learning and affinity-learning, are equipped with pixel-level semantic correlation in most cases. The former constructs the class-agnostic prior mask by maximizing pair-wise pixel similarity to compensate for the missing information caused by feature compression~\citep{wang2019panet,li2021adaptive,tian2022prior,lang2024few}, while the latter directly leverages the similarity of support-query pixel pairs to aggregate support semantics into the query feature~\citep{zhang2021few,wang2022adaptive,zhang2022feature,xu2023self}. For instance, CyCTR~\citep{zhang2021few} introduced a cycle-consistent transformer to selectively aggregate support-pixel semantics and SCCAN~\citep{xu2023self} designed a self-calibrated cross-attention to alleviate the pixel-mismatch issue.

\begin{figure}[t]
\setlength{\abovecaptionskip}{1pt}
\centering
\includegraphics[width=1.0\linewidth]{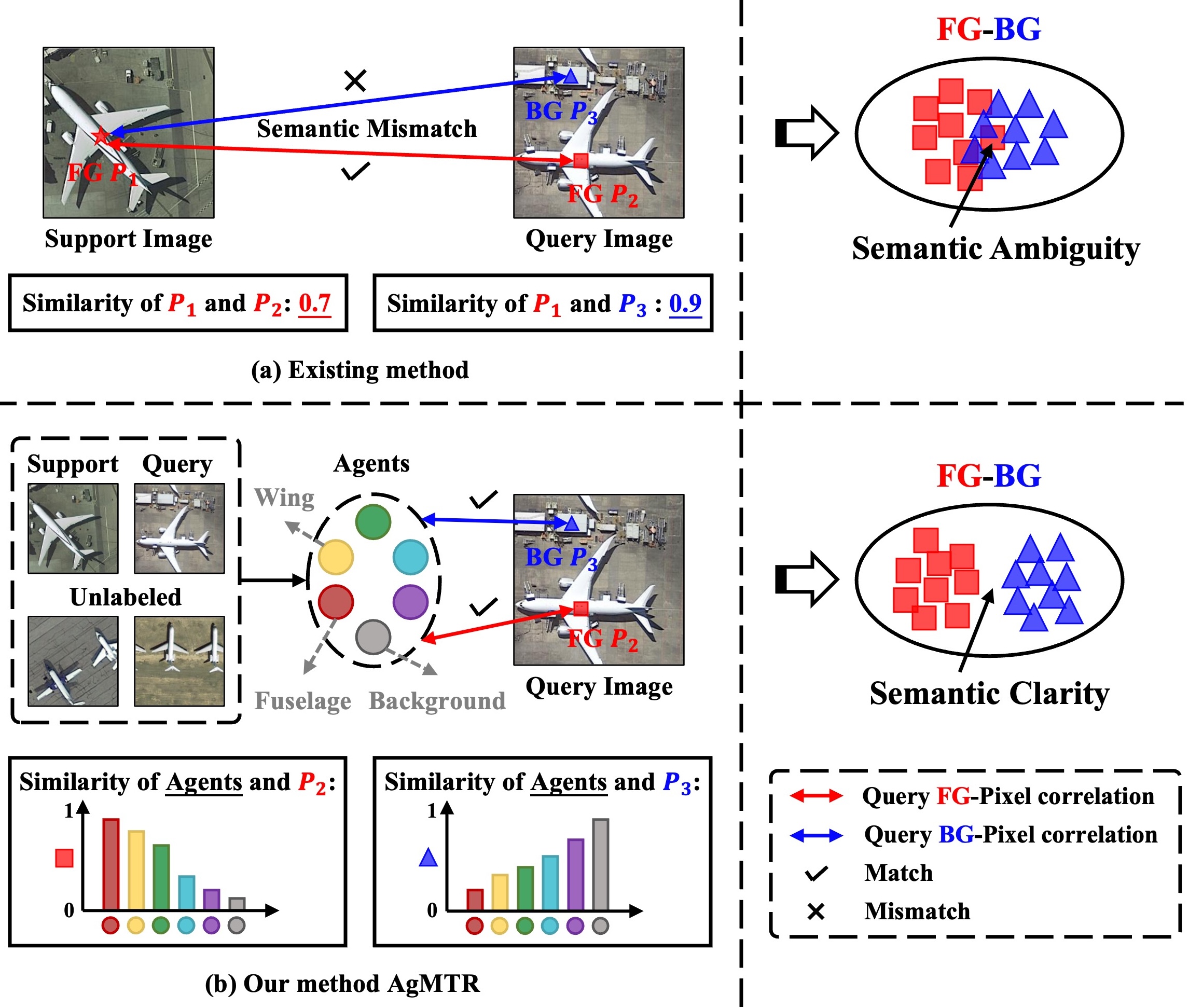}
\caption{ 
(a) Due to the extreme intra-class variations and cluttered backgrounds in remote sensing scenes, directly leveraging the correlations between support-query pixel pairs to aggregate support semantics may result in tremendous mismatches. In this case, both FG and BG pixels of the query are likely to aggregate the FG semantics of the support, resulting in semantic ambiguity. (b) AgMTR mines a set of local-aware agents from support, unlabeled, and query images for constructing agent-level semantic correlation. At this point, query FG-pixel $P_2$ will selectively aggregate the agent semantics responsible for the `Fuselage', while the BG-pixel $P_3$ will aggregate the background agent semantics, implementing semantic clarity between the query FG and BG pixels.}
\label{fig:1}
\end{figure}

\begin{figure*}[t]
\setlength{\abovecaptionskip}{1pt}
\centering
\includegraphics[width=1.0\linewidth]{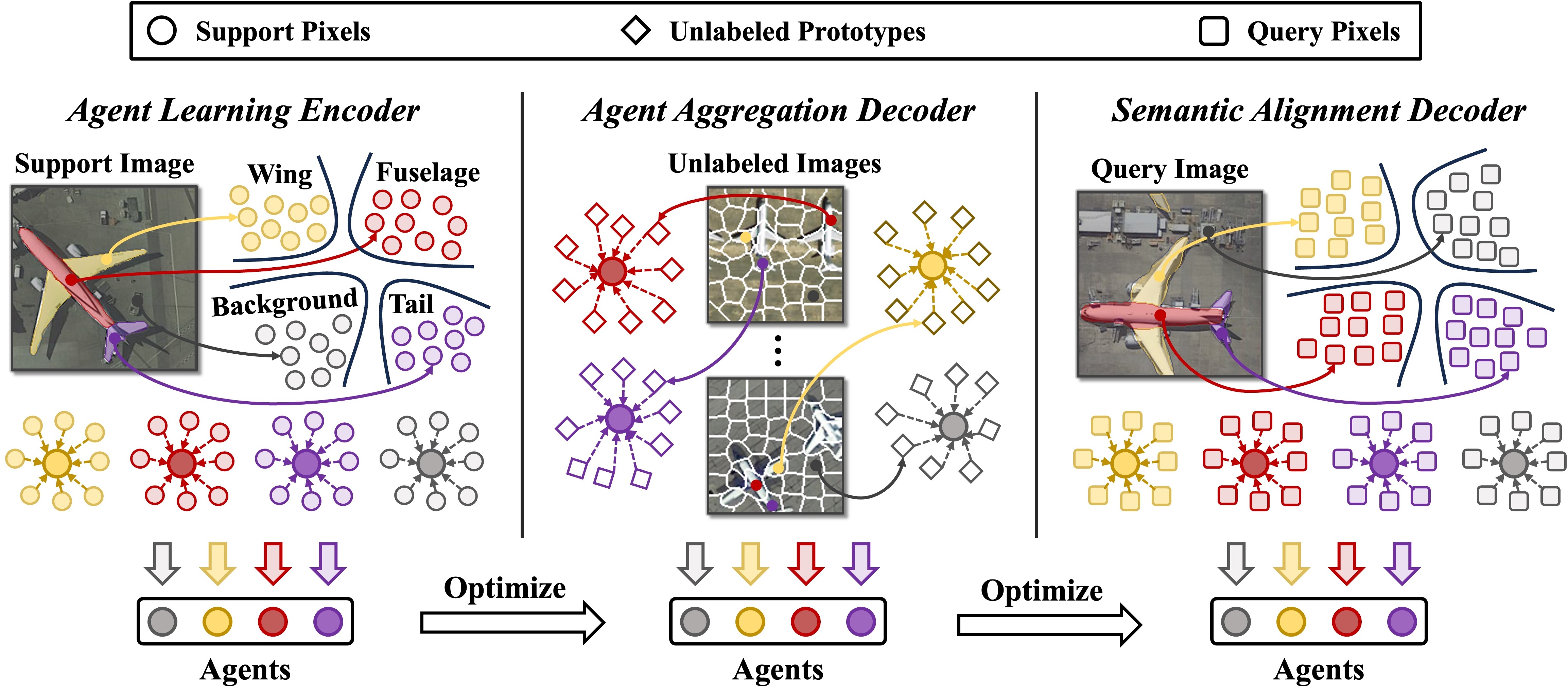}
\caption{\textbf{The specific process of mining agents.} Firstly, Agent Learning Encoder (ALE) dynamically divides the image into different but complementary local regions, e.g., `Fuselage', `Wing', `Tail', `Background', etc., via the support foreground mask, thus assigning different local semantics to different agents. To further optimize the agents, the Agent Aggregation Decoder (AAD) introduces a set of unlabeled images and obtains a set of local prototypes through unsupervised clustering, where the agents will selectively perceive and aggregate fine-grained information from these prototypes. Finally, the Semantic Alignment Decoder (SAD) constructs the query pseudo-local masks to create conditions for the agents to aggregate the local semantics from the query image.}
\label{fig:1_1}
\end{figure*}

Despite the desirable results in natural scenarios, when extended to remote sensing scenarios with extreme intra-class variations and cluttered backgrounds, directly leveraging the correlations between support-query pixel pairs to aggregate support semantics may lead to tremendous mismatches. Unfortunately, the low-data regime of FSS would exacerbate these negative effects. As illustrated in Fig.\ref{fig:1}(a), in view of the appearance difference between objects in the support-query pair and the complex background interference, the background (BG) pixel `Roof' $P_3$ in the query even matches the foreground (FG) pixel `Plane' $P_1$ in the support more than the FG-Pixel `Plane' $P_2$ in the query. At this point, both query FG and BG pixels will aggregate support FG semantics, resulting in semantic ambiguity between the FG and BG pixels of the query and yielding segmentation failures. 

To correct the semantic ambiguity between the query FG and BG pixels caused by pixel-level mismatch, we suggest adaptively mining a set of local-aware agents (i.e., condensed feature representations) for constructing agent-level semantic correlation. Note that different agents will highlight different local regions of the interested class, e.g., the `Fuselage', `Wings', etc., of the `Plane'. In this case, different pixels in the query image will derive the corresponding agent-level semantic correlations, thus executing specific semantic aggregations. As illustrated in Fig.\ref{fig:1}(b), given query FG-pixel $P_2$ selectively aggregates the agent semantics responsible for the `Fuselage', while the BG-pixel $P_3$ aggregates the background agent semantics. This is due to the fact that, compared to pixel-level semantics, the agents are equipped with local contextual information and possess a broader receptive field, thus ensuring the safe execution of semantic aggregation for query pixels and enhancing semantic clarity between query FG and BG pixels.

Nevertheless, excavating representative agents without explicit supervision is not trivial. This paper develops an Agent Mining Transformer network called \textbf{AgMTR} to address this issue. For minimizing the negative impact of large intra-class variations in remote sensing scenarios, besides routinely exploiting the \textbf{support images}, AgMTR focuses on exploring valuable semantics from \textbf{unlabeled images} and the \textbf{query image} itself to optimize agents. The specific process of mining agents is depicted in Fig.\ref{fig:1_1}.

Firstly, the \textbf{Agent Learning Encoder (ALE)} is proposed to guide agents to efficiently mine salient semantics (i.e., class-specific semantics) from the support pixels that can help segment the target classes in the query image in a masked cross-attention manner~\citep{cheng2022masked}. To enhance the semantic diversity of the agents, ALE dynamically imposes equal division constraints on the foreground region, thus assigning different yet complementary local masks to different agents. At this point, different agents can aggregate the pixel semantics from the corresponding local regions, achieving the local-awareness of the target category, such as the `Fuselage', and `Wings' of the `Plane'.


Merely aggregating support semantics is not enough to bridge the semantic gap from the image pair. The \textbf{Agent Aggregation Decoder (AAD)} is proposed to introduce unlabeled images containing the interested class as references, with the expectation that agents can adaptively explore the corresponding object semantics from unlabeled data sources, thus breaking beyond the limited support set for better guiding the query segmentation. Notably, the unlabeled image features will be adaptively clustered into a set of local prototypes, where the proposed agents will selectively aggregate these prototype semantics to further enhance the local-awareness of the target category.

Furthermore, inspired by the strong intra-object similarity, i.e., pixels within the one object are more similar than pixels between different objects~\citep{wang2024focus}, learning from the query image itself is equally crucial to guide its own segmentation. The \textbf{Semantic Alignment Decoder (SAD)} is suggested to promote semantic consistency between the agents and the query object. Specifically, the pseudo-local masks of the query image will be constructed by computing the similarity between the different local agents and the query feature. Under the constraint of the local masks, different agents will further aggregate the corresponding query local semantics, thus aligning with the query semantics.

Based on the above three components, AgMTR constructs precise agent-level semantic correlations, effectively distinguishing FG and BG pixels in the semantic aggregation of the query feature. 

The primary contributions can be summarized as follows:

\begin{itemize}
\item A novel Agent Mining Transformer (AgMTR) network for FSS in remote sensing is proposed to adaptively mine a set of local-aware agents for different objects to construct agent-level semantic correlation, thus correcting the semantic ambiguity between the query FG and BG pixels caused by pixel-level mismatches. 
\item We construct an Agent Learning Encoder, Agent Aggregation Decoder, and Semantic Alignment Decoder to adaptively mine and optimize agent semantics from support, unlabeled, and query images, in that order.
\item Extensive experiments on remote sensing FSS benchmark iSAID, validate the superiority of the proposed AgMTR and achieve state-of-the-art performance. Surprisingly, our method likewise outperforms previous SOTA on two popular CV benchmarks, PASCAL-5$^i$, and COCO-20$^i$.
\end{itemize}

\section{Related work}\label{Related work}
\subsection{Semantic Segmentation}
Semantic segmentation, with the goal of predicting pixel-level labels within an image, has been widely emphasized by researchers as a fundamental task in image interpretation. Long et al.~\citep{long2015fully} first proposed a full convolutional network (FCN) for the semantic segmentation task, providing a solid foundation for the following research. To further mine the association of contextual information, Chen et al.~\citep{chen2014semantic} proposed Dilated Convolution to expand the convolutional receptive field. Meanwhile, Zhao et al.~\citep{zhao2017pyramid} and Chen et al.~\citep{chen2017deeplab} proposed Pyramid Pooling Module (PPM) and Atrous Spatial Pyramid Pooling (ASPP), respectively, to combine the feature representations between different scales and regions for multi-scale semantic aggregation. 

On this basis, some researchers have focused on designing semantic segmentation methods specific to remote sensing scenes. Feng et al.~\citep{feng2020npaloss} proposed a neighboring pixel affinity loss (NPALoss) to focus on the optimization for small-size objects and hard object boundaries. Peng et al.~\citep{peng2021cross} proposed a Cross Fusion Network (CF-Net) for fast and effective extraction of small-scale objects. In addition, Niu et al.~\citep{9580861} decoupled the semantic segmentation task and proposed a disentangled learning paradigm to model foreground and boundary objects separately. Despite promising success with large-scale data, these methods fail to perform well when generalizing to unseen domains.

\subsection{Few-shot Learning}
Recently community researchers have proposed Few-shot Learning (FSL)~\citep{xu2024attention,finn2017model,jamal2019task,snell2017prototypical} for fast generalization to unseen classes, which aims to recognize novel (unseen) classes from a handful of labeled samples~\citep{wang2020generalizing}. In general, FSL methods would employ the meta-learning paradigm~\citep{finn2017model}, where the model understands representative knowledge from the training dataset (seen classes) and generalizes it to novel tasks (unseen classes). On this basis, these methods can be divided into the following two branches: (i) Optimization-based method~\citep{finn2017model,li2017meta,jamal2019task,sun2019meta}, which aims to search for a set of proper parameters for the model that can be easily generalized to various novel tasks. For example, Finn et al.~\citep{finn2017model} learned the initialization parameters of the model while Ravi et al.~\citep{ravi2016optimization} learned an optimization strategy instead of a fixed optimizer to facilitate faster convergence. (ii) Metric-based method~\citep{vinyals2016matching,snell2017prototypical,sung2018learning,ye2020few,zhang2022deepemd}, which aims to find a specific embedding space that can efficiently distinguish different classes for classification. For example, Zhang et al.~\citep{zhang2022deepemd} utilized Earth Mover's Distance (EMD) to model the distance between images as the optimal transport plan. Bateni et al.~\citep{bateni2020improved} proposed to construct Mahalanobis-distance classifiers to improve the accuracy of few-shot classification. 

In addition, given the large intra-class variations and small inter-class variations of remote sensing scenarios, some FSL methods specific to remote sensing scenarios have been successively proposed~\citep{alajaji2020few,zhong2020few,cheng2021spnet,kim2021saffnet,zhu2023hcpnet}. In this paper, we instead apply FSL to the more advanced task of scene understanding rather than scene classification in remote sensing scenes, i.e., semantic segmentation.

\subsection{Few-shot Segmentation}\label{Few-shot segmentation}
Few-shot Segmentation (FSS)~\citep{wang2019panet,zhang2019canet,bi2024prompt,liu2022learning,zhang2021self}, a natural extension of Few-shot Learning to tackle intensive prediction tasks, is proposed to guide the model to activate corresponding objects in the query image utilizing several labeled samples (i.e., support images) containing the interested class. Current FSS methods can be broadly divided into two parts: (i) Prototype-learning, which aims to compress the support feature into a prototype through masked average pooling (MAP), and perform feature comparison with the query feature to segment the interested object~\citep{wang2019panet,zhang2019canet,liu2022learning,zhang2021self}. However, these methods would inevitably lose spatial information and lack context-awareness. To address this issue, besides attempting to generate multiple prototypes~\citep{liu2020part,yang2020prototype}, they typically constructed the class-agnostic prior mask by maximizing pair-wise pixel similarity to further mine pixel-level support semantics~\citep{wang2019panet,li2021adaptive,tian2022prior,lang2024few}. (ii) Affinity-learning~\citep{zhang2021few,wang2022adaptive,zhang2022feature,xu2023self}, which aims to directly construct pixel-level feature matching for segmentation. For instance, Zhang et al.~\citep{zhang2021few} introduced a cycle-consistent transformer to aggregate support-pixels semantics selectively. Xu et al.~\citep{xu2023self} designed a self-calibrated cross-attention to alleviate the pixel-mismatch issue. However, such pixel-level similarity would ignore contextual information and tend to be affected by background clutter or noisy pixels causing semantic ambiguity. We propose feature embeddings called agents, which are condensed feature representations similar to prototypes, and at the same time mine the valuable pixel semantics for dynamic optimization, which can combine the strengths of prototype learning and affinity learning.

In addition, several methods, such as BAM~\citep{lang2023base}, D$^2$Zero~\citep{he2023semantic}, and PADing~\citep{he2023primitive}, focus on addressing the bias toward seen classes in Few-shot (Zero-shot) scenes for better generalize to unseen classes. In the field of remote sensing, the Few-shot Segmentation task has also received an enthusiastic response~\citep{yao2021scale,wang2021dmml,jiang2022few,lang2023global,10152484,bi2023not}. Yao et al.~\citep{yao2021scale} proposed a scale-aware prototype matching scheme to handle the large variance of objects' appearances and scales. Bi et al.~\citep{bi2023not} designed DMNet to mine the class-specific semantics from query images. To maximize the breakthrough of the few-shot limitations for better adapting to remote sensing scenarios, this paper proposes to mine and optimize class-specific semantics from support, unlabeled, and query images.

\begin{figure*}[t]
\setlength{\abovecaptionskip}{1pt} \setlength{\belowcaptionskip}{1pt}
\centering
\includegraphics[width=1.0\linewidth]{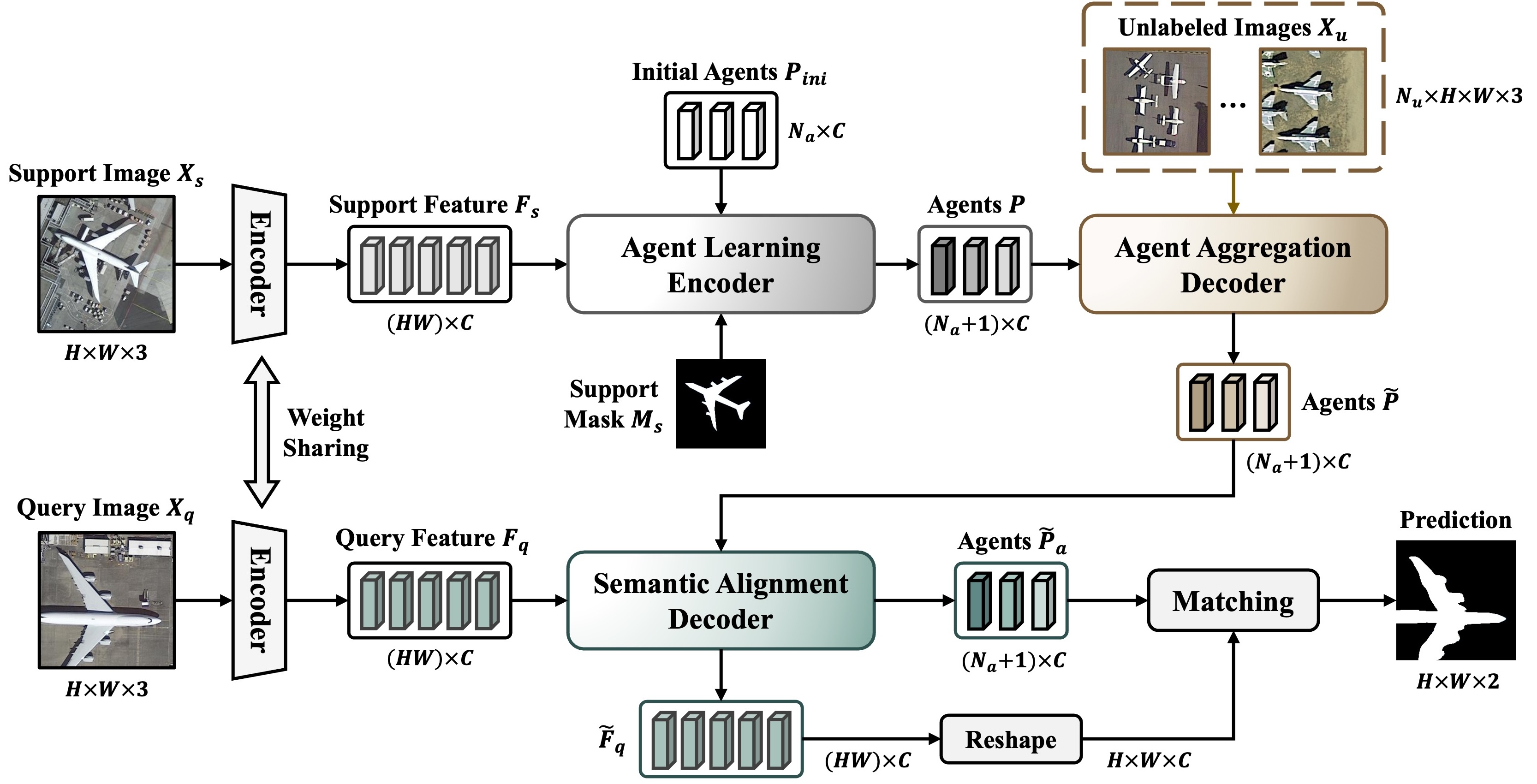}
\caption{\textbf{The overall pipeline of the proposed AgMTR,} aims at constructing agent-level semantic correlation to correct the semantic ambiguity between the query FG and BG pixels caused by pixel-level mismatches. Three components are proposed to excavate representative agents without explicit supervision: Agent Learning Encoder (ALE), Agent Aggregation Decoder (AAD), and Semantic Alignment Decoder (SAD). First, ALE dynamically divides the support mask into multiple local masks, thus guiding different agents to mine the semantics under different local masks for local-awareness. AAD then introduces a series of unlabeled images where the agents could selectively perceive and aggregate semantics beneficial to them from the unlabeled data source to break through the limited support set. Finally, SAD aims to mine the query image's own interested semantics to promote semantic consistency between the agents and the query object.}
\label{fig:2}
\end{figure*}

\section{Problem Definition}\label{Problem Definition}
The purpose of Few-shot Segmentation (FSS) is to segment unseen class objects without additional training by utilizing only a handful of labeled samples. Following the mainstream schemes~\citep{bi2023not,lang2024few}, the meta-learning paradigm is employed to train the model. Concretely, given a training dataset $D_{train}$ with seen classes $C_{seen}$ and a testing dataset $D_{test}$ with unseen classes $C_{unseen}$ ($ C_{seen} \cap C_{unseen}= \emptyset $), the model would understand representative knowledge from multiple tasks (i.e., episodes) sampled from $D_{train}$, and then generalizes directly to $D_{test}$. Specifically, both datasets $D_{train}$ and $D_{test}$ consist of multiple episodes, each containing a support set $ S =\left \{ \left ( X^i_s, M^i_s\right )  \right \}_{i=1}^{K} $  and a query set $ Q =\left \{ \left ( X_q, M_q\right )  \right \} $, where $X_{*}^{i}$ and $M_{*}^{i}$ denote the image and the corresponding mask containing the interested class, respectively, $K$ denotes the number of support image provided in $S$. During each training episode, the FSS models are optimized by predicting the query image $X_q$ with the guidance of the support set $S$. After training, the model evaluates the performance in each testing episode on $D_{test}$ without any optimization. Note that the query mask $M_{q}$ is not available during the testing phase.

Variant: Considering that extreme intra-class variations in remote sensing scenarios would further exacerbate the gap between support-query image pairs in FSS, it is suboptimal and insufficient for previous work to utilize only support images (i.e., labeled samples) to guide the query image. Thus, this paper also introduces unlabeled sample set $S_u =\left \{ \left ( X^i_u\right )  \right \}_{i=1}^{N_u} $ with the same interested class as the labeled samples to assist in segmentation. These samples are derived by random sampling from the interested class during the training and testing phase, with the corresponding masks removed. At this point, the model will segment the image $X_q$ with the guidance of $S$ and $S_u$.

\begin{figure*}[t]
\setlength{\abovecaptionskip}{1pt} \setlength{\belowcaptionskip}{1pt}
\centering
\includegraphics[width=0.65\linewidth]{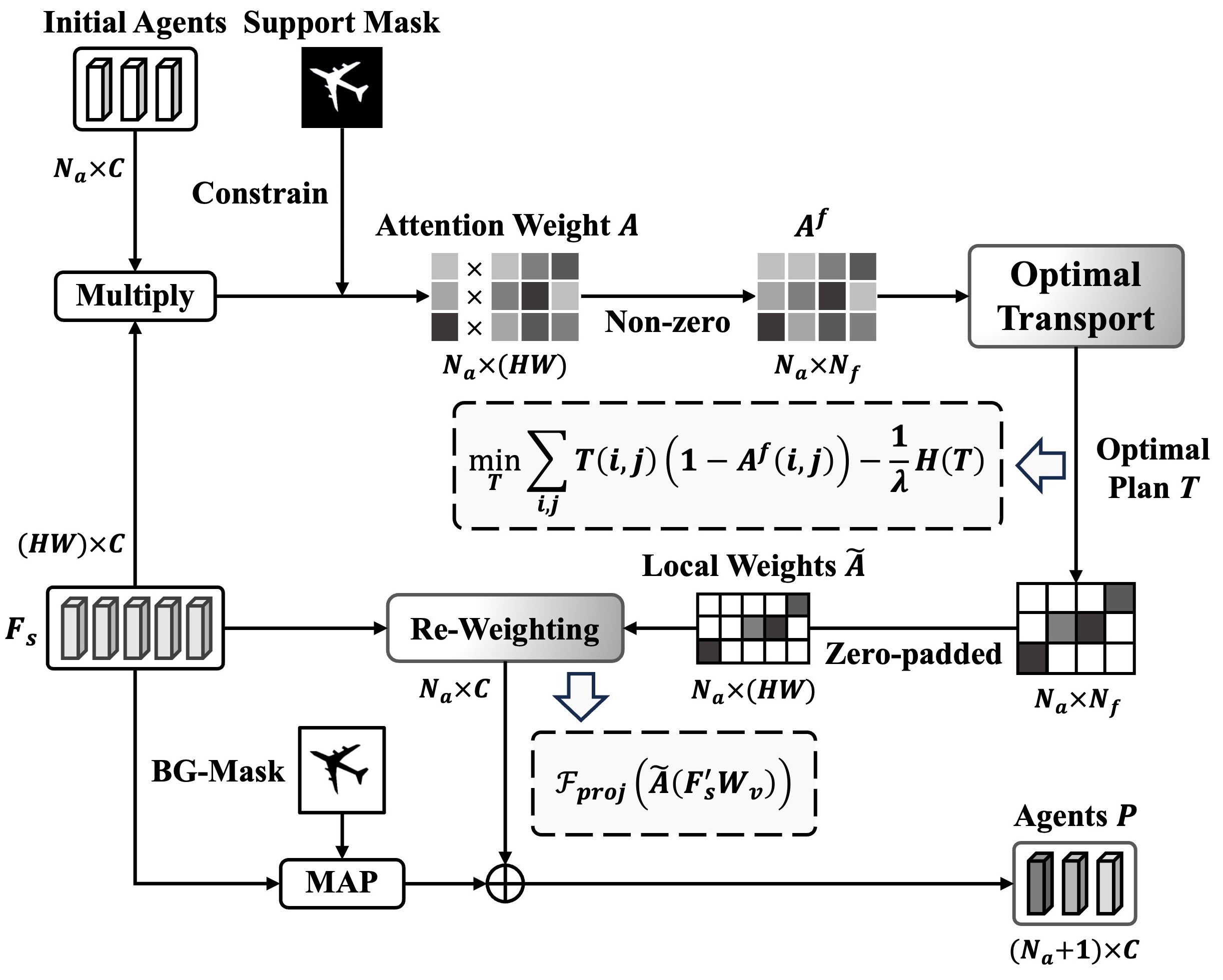}
\caption{\textbf{The detailed pipeline of Agent Learning Encoder (ALE),} which aims at driving the agents to efficiently capture interested information from support pixels through the masked cross-attention mechanism. To enhance the semantic diversity of agents, ALE dynamically decomposes the foreground mask into different yet complementary local masks, thus ensuring that different agents aggregate different local semantics.}
\label{fig:3}
\end{figure*}

\section{Proposed Method}\label{Proposed Method}
\subsection{Method Overview}

To correct the semantic ambiguity of query FG and BG pixels caused by the semantic mismatch in previous pixel-level semantic correlation, this paper develops an Agent Mining Transformer network named AgMTR, which suggests mining a set of local-aware agents (i.e., condensed feature representations similar to prototypes), and relies on them to construct agent-level semantic correlation. Across tasks, the proposed agents are not static, or an agent contains the local semantics of multiple classes. On the contrary, the semantic awareness of the agents is dynamic, where AgMTR would adaptively mine the corresponding class-specific semantics according to the target classes of different episodes/tasks, thus generating local-aware agents for the current class. Compared to prototypes, agents offer more precise and stable feature representations, a richer and more diverse semantic source, and enhanced local contextual awareness. 

The overall pipeline is depicted in Fig.\ref{fig:2}. In particular, given the support-query image pair $X_s$ and $X_q$, we extract features with a shared-feature encoder (e.g., ViT~\citep{dosovitskiy2020image} and DeiT~\citep{touvron2021training}) to derive the support feature $F_s\in \mathbb{R}{^{H \times W  \times C}}$ and query feature $F_q\in \mathbb{R}{^{H \times W  \times C}}$. For more matched agent-level semantic correlation, AgMTR constructs three components, Agent Learning Encoder (ALE), Agent Aggregation Decoder (AAD), and Semantic Alignment Decoder (SAD), to dynamically mine and optimize agent semantics from support, unlabeled, and query images, in that order:
\vspace{-2pt}
\begin{itemize}
\item \textbf{Agent Learning Encoder:} ALE dynamically decomposes the support mask into multiple local masks and arranges different initial agents to aggregate the support semantics under different local masks for deriving local-awareness of the target category.
\item \textbf{Agent Aggregation Decoder:} To break through the limited support set for better guidance, AAD introduces unlabeled images containing the interested class, where the agents will selectively perceive and aggregate the semantics beneficial to them from unlabeled images.
\item \textbf{Semantic Alignment Decoder:} To further bridge the semantic gap, SAD constructs the pseudo-local masks of the query image, which drives the agents to aggregate the local semantics of the query itself to promote semantic consistency with the query object.
\end{itemize}

\subsection{Agent Learning Encoder} \label{sec:Agent Learning Encoder}
As mentioned in Sec.\ref{Indroduction}, we suggest mining a set of context-rich agents to perform agent-level semantic correlation replacing the previous pixel-level. At this point, the focus turns to capturing representative agents. Agent Learning Encoder aims to guide agents to efficiently mine salient semantics (i.e., class-specific semantics) from the support pixels that can help segment the target classes in the query image, which is the most basic and important step, with the detailed pipeline illustrated in Algorithm \ref{Agent Learning Encoder}.

\noindent \textbf{Agent Initialization:} Firstly, as depicted in Fig.\ref{fig:3}, the initial agents $P^{ini}=\left [ p^{ini}_1, p^{ini}_2,...,p^{ini}_{N_a} \right ]  \in \mathbb{R}{^{N_a \times C}}$ are derived by adding between randomly initialized feature embeddings $\left \{t_i  \right \}_1^{N_a} \in \mathbb{R}{^{N_a \times C}}$ and the support prototype obtained by performing masked average pooling (MAP) operation:
\begin{align}  \label{equation:1}
\setlength{\abovecaptionskip}{1pt}
\setlength{\belowcaptionskip}{1pt}
p_i^{ini}=\mathrm{MAP}(F_s\odot M_s)+t_i,i=1,2,...,N_a
\end{align}
where $\mathrm{MAP}\left (\cdot\right )$ denotes the compression of the support feature $F_s$ under the foreground mask $M_s$ into a feature vector and $\odot$ denotes the dot product. Notably, the randomly initialized feature embeddings $\left \{t_i  \right \}_1^{N_a}$ obeys a normal distribution.

\begin{algorithm}[t]
\caption{ Agent Learning Encoder (ALE) }\label{Agent Learning Encoder}
\KwIn{ Support feature $F_s$, Support mask $M_s$}
\KwOut{ Local-aware agents $P$}

\textbf{Step 1: Agent Initialization}

Generate initial agents $P^{ini} \leftarrow \left( F_s, M_s \right)$ according to Eq.(\ref{equation:1});

\textbf{Step 2: Agent Learning}

Construct attention weight between agents and the feature $A \leftarrow \left( P^{ini}, F_s, M_s \right)$ according to Eq.(\ref{equation:2})-Eq.(\ref{equation:3});

Decompose the attention weight $A$ into different but complementary local attention weights $\tilde{A}$ according to Eq.(\ref{equation:4})-Eq.(\ref{equation:5});

Generate corresponding local agent based on different local attention weights $P \leftarrow \left(\tilde{A},F_s \right)$ according to Eq.(\ref{equation:6});
\end{algorithm}

\noindent \textbf{Agent Learning:} To empower agents with context-awareness, ALE employs masked cross-attention~\citep{cheng2022masked} to aggregate the interested semantics in the support feature to corresponding agents. At this point, the attention weight $A \in \mathbb{R}{^{N_a \times HW}}$between agents and feature can be formulated as:
\begin{align}  \label{equation:2}
\setlength{\abovecaptionskip}{1pt}
\setlength{\belowcaptionskip}{1pt}
A=\mathrm{softmax}\left ( \frac{\left (P^{ini}W_q  \right )  \left (F_s^{'}W_k  \right )^T}{\sqrt{d}}+\tilde{M}_s   \right ) 
\end{align}
where $W_*\in \mathbb{R}{^{C \times C}}$ denote projection weights, $F_s^{'}\in \mathbb{R}{^{HW \times C}}$ denotes the flattened support feature. Notably, we utilize $\tilde{M}_s$ to constrain the region of interest for the attention weight:
\begin{align}  \label{equation:3}
\setlength{\abovecaptionskip}{1pt}
\setlength{\belowcaptionskip}{1pt}
\tilde{M}_s\left ( i \right )=\begin{cases}0,\quad \mathrm{if}\ \ M_s^{'}\left ( i \right ) =1
\\
-\infty,\quad \mathrm{otherwise}
\end{cases}
\end{align}
where $M_s^{'}$ denotes the flattened support mask. After normalization in Eq.(\ref{equation:2}), the attention weight $A$ focuses only on the foreground region of the support mask, thus ensuring that the agents aggregate only the interested foreground semantics.

Without supervising the learning process, such agents tend to capture similar foreground semantics resulting in redundancy. To enhance the semantic diversity of agents, we propose decomposing the attention weight into different but complementary local attention weights, i.e., decomposing the support mask into different local masks. Under the constraint of those local masks, different agents can aggregate the semantics from different local regions, yielding different local perceptions of the target category.

Concretely, we model this allocation as the Optimal Transport (OT) problem. By searching for the transport plan with the minimum transport cost, the pixel allocation of different agents is accomplished, which can be solved efficiently by the Sinkhorn~\citep{cuturi2013sinkhorn} algorithm. We define the cost matrix as $(1-A^f)$, where $A^f \in \mathbb{R}{^{N_a \times N_f}}$ denotes the similarity matrix between the agents and the foreground feature of the support image instead of the whole feature. Notably, the larger $A^f(i,j)$ is, the greater the similarity between the $i^{th}$ agent and the $j^{th}$ support pixel, the lower the transport cost, and the greater the likelihood that the pixel will be assigned to that agent. At this point, the transport plan is denoted as $T\in \mathbb{R}{^{N_a \times N_f}}$, with the optimization function formulated as:
\begin{align}  \label{equation:4}
\setlength{\abovecaptionskip}{1pt}
\setlength{\belowcaptionskip}{1pt}
\min_{T\in \Upsilon }\sum_{i,j}T\left ( i,j \right )\left ( 1-A^f\left ( i,j \right )  \right ) -\frac{1}{\lambda } H\left ( T \right )
\end{align}
where $H\left ( T \right )=-\sum_{i,j}T(i,j)\log T(i,j)$ denotes entropy regularization and $\lambda$ is responsible for adjusting the degree of entropy impact. Besides, the transport plan $T$ needs to obey the distribution of $\Upsilon$:

\begin{align}  \label{equation:5}
\setlength{\abovecaptionskip}{1pt}
\setlength{\belowcaptionskip}{1pt}
\Upsilon =\left \{ T \in \mathbb{R}{^{N_a \times N_f}}| T\mathbf{I}_{m}=\frac{1}{N_a} \mathbf{I}_{n} ,T^T\mathbf{I}_{n}=\frac{1}{N_f} \mathbf{I}_{m}  \right \} 
\end{align}
where $\mathbf{I_m}\in \mathbb{R}{^{N_f}}$ and $\mathbf{I_n}\in \mathbb{R}{^{N_a}}$ denote vectors with values of 1 in all dimensions. Eq.(\ref{equation:5}) aims to force each agent to allocate the same number of foreground pixels, avoiding a situation where all pixels are allocated to one agent. 

After deriving the optimal transport plan $T$, we zero-padded it to recover the background region to produce the local attention weights $\tilde{A}\in \mathbb{R}{^{N_a \times HW}}$. At this point, different agents $P$ will aggregate the corresponding pixel semantics under the constraint of local weights:
\begin{align}  \label{equation:6}
\setlength{\abovecaptionskip}{1pt}
\setlength{\belowcaptionskip}{1pt}
P=\mathcal{F}_{proj}\left ( \tilde{A}\left (F_s^{'}W_v\right )   \right ) 
\end{align}
where $\mathcal{F}_{proj}$ denotes the two-layer MLP and $W_v$ denotes the projection weight. 

Note that the agents are not the multiple local regions divided by the ALE, but rather the product of condensing the pixel semantics of different local regions into a single feature vector. Finally, the background prototype derived by performing MAP on the support feature will be incorporated as the background agent in $P= \left [ p_1,p_2,...,p_{N_a}, p_{N_a+1}\right ] \in \mathbb{R}{^{(N_a+1 )\times C}}$.

\subsection{Agent Aggregation Decoder}\label{sec: Agent Aggregation Decoder}

Capturing the class-specific semantics from the support images merely is not sufficient to bridge the semantic gap between the image pairs, especially in complex remote sensing scenarios. To break through the limited support set for mitigating the intra-class variations, the Agent Aggregation Decoder introduces unlabeled images containing the interested class as a reference, expecting agents to adaptively explore the richer interested semantics from the unlabeled data source, with the pipeline illustrated in Algorithm \ref{Agent Aggregation Decoder}. 

\begin{figure*}[t]
\setlength{\abovecaptionskip}{1pt} \setlength{\belowcaptionskip}{1pt}
\centering
\includegraphics[width=0.65\linewidth]{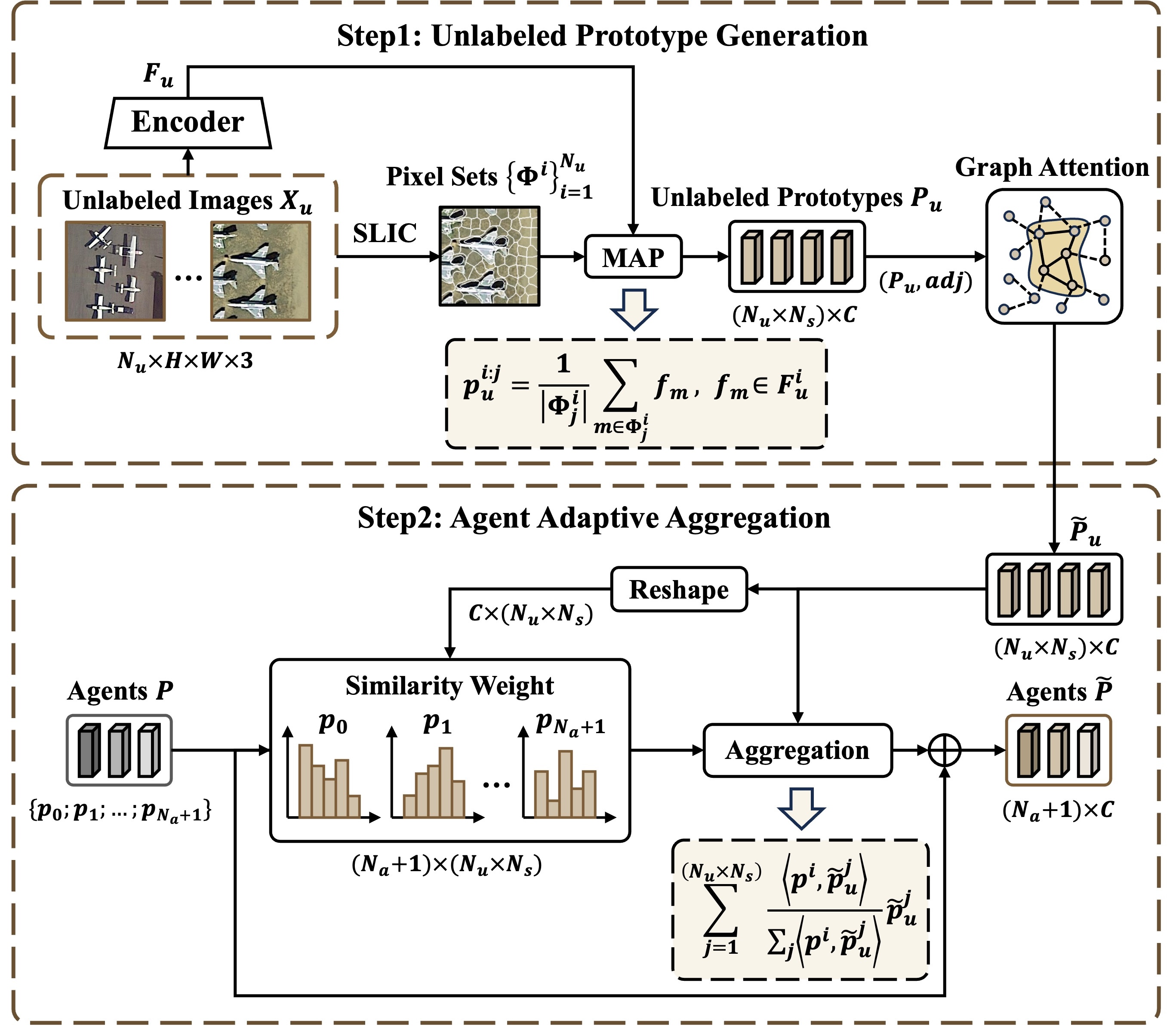}
\caption{\textbf{The detailed pipeline of Agent Aggregation Decoder (AAD).} AAD introduces unlabeled images containing the interested class as a reference, expecting agents to adaptively explore the richer interested semantics from unlabeled data sources to break through the support set limitations.}
\label{fig:4}
\end{figure*}

\begin{algorithm}[t]
\caption{Agent Aggregation Decoder (AAD) }\label{Agent Aggregation Decoder}
\KwIn{ Local agents $P$, Unlabeled sample set $S_u$}
\KwOut{ Refined local agents $\widetilde{P}$}

\textbf{Step 1: Unlabeled prototype generation}

First, generate unlabeled features $F_u \leftarrow S_u$;

Obtain a series of pixel sets by clustering and compress them into local prototypes $P_u \leftarrow F_u$ according to Eq.(\ref{equation:7});

Enhance the contextual semantics of the unlabeled prototypes $\widetilde{P}_u \leftarrow P_u$ according to Eq.(\ref{equation:8});

\textbf{Step 2: Agent Adaptive Aggregation}

Different agents adaptively aggregate the beneficial semantics for them from unlabeled prototypes $\widetilde{P} \leftarrow \left( P, \widetilde{P}_u \right)$ according to Eq.(\ref{equation:9});

\end{algorithm}

\noindent \textbf{Unlabeled prototype generation:} Specifically, as illustrated in Fig.\ref{fig:4}, given $N_u$ unlabeled images $X_u$, we extract features via a shared encoder to generate unlabeled features $F_u \in \mathbb{R}{^{N_u\times H \times W  \times C}}$. Considering that there is a significant amount of background noise in the unlabeled feature that cannot be directly exploited, we employ a super-pixel scheme SLIC~\citep{achanta2012slic} to cluster the unlabeled image into $N_s$ pixel sets $\Phi^i =\left \{ \Phi _1^i,\Phi _2^i,...,\Phi _{N_s}^i \right \}, i=1,2,..., N_u $ based on color, texture, and distance. The unlabeled features can be compressed into a series of local prototypes $P_u\in \mathbb{R}{^{(N_u\times N_s) \times C}}$ through MAP operation:
\begin{align}  \label{equation:7}
\setlength{\abovecaptionskip}{1pt}
\setlength{\belowcaptionskip}{1pt}
p_u^{i;j}=\frac{1}{\left | \Phi _j^i \right | } \sum_{m\in\Phi _j^i}f_m, \quad f_m\in F_u^i
\end{align}
where $p_u^{i;j} \in \mathbb{R}{^{C}}$ denotes the $j^{th}$ local prototype generated by the $i^{th}$ unlabeled image and $\Phi _j^i$ denotes the $j^{th}$ pixel set generated by the $i^{th}$ unlabeled image. 

To further enhance the contextual semantics of the unlabeled prototype set, we model it as the graph, with the prototypes considered as nodes in the graph, employing a graph attention network (GAT)~\citep{velivckovic2017graph} to enhance the semantic associations between nodes, which can be formulated as:
\begin{equation}\label{equation:8}
\setlength{\abovecaptionskip}{1pt}
\setlength{\belowcaptionskip}{1pt}
\begin{split}
&\widetilde{P}_u =\mathrm{GAT}\left ( P_u,adj \right )  \\
&s.t.\quad adj\left(m,n\right)=\begin{cases}1,\quad \mathrm{if} \ \ \left \langle p_{u}^{m},p_{u}^{n} \right \rangle>\delta 
\\
0,\quad \mathrm{otherwise}
\end{cases}
\end{split}
\end{equation}
where $\mathrm{GAT}$ denotes the graph attention network, $adj\in \mathbb{R}{^{(N_u\times N_s) \times (N_u\times N_s)}}$ denotes the adjacency matrix formed between the nodes and $\left \langle \cdot,\cdot \right \rangle$ denotes the cosine similarity function, $p_u^{*} \in \mathbb{R}{^{C}}$ denotes the $*^{th}$ prototype in the unlabeled prototype set, and $\delta$ is a threshold to determine whether the nodes are neighboring or not, which is set to 0.5 in the experiments.

\noindent \textbf{Agent Adaptive Aggregation:} Given the augmented prototypes of unlabeled features $\widetilde{P}_u \in \mathbb{R}{^{(N_u\times N_s) \times C}}$, the agents $P \in \mathbb{R}{^{(N_a+1) \times C}}$ will selectively perceive and aggregate object semantics from these prototypes that are beneficial to them in a cross-attention manner. As depicted in Fig.\ref{fig:4}, by performing similarity computation with unlabeled prototypes $\widetilde{P}_u$, different agents $P$ will receive different similarity weights, resulting in different weighted semantic aggregations:

\begin{align}  \label{equation:9}
\setlength{\abovecaptionskip}{1pt}
\setlength{\belowcaptionskip}{1pt}
\tilde{p}^i=p^i+\beta \sum_{j=1}^{(N_u\times N_s)} \frac{\left \langle p^i,\tilde{p}_u^j \right \rangle  }{\sum_{j}\left \langle p^i,\tilde{p}_u^j \right \rangle   } \tilde{p}_u^j
\end{align}
where $\beta$ denotes the learnable weight to balance the semantic aggregation, which is initialized to 0.4. At this point, the local agents $\widetilde{P}\in \mathbb{R}{^{(N_a+1) \times C}}$ will be further optimized.

\begin{figure*}[t]
\setlength{\abovecaptionskip}{1pt} \setlength{\belowcaptionskip}{1pt}
\centering
\includegraphics[width=0.65\linewidth]{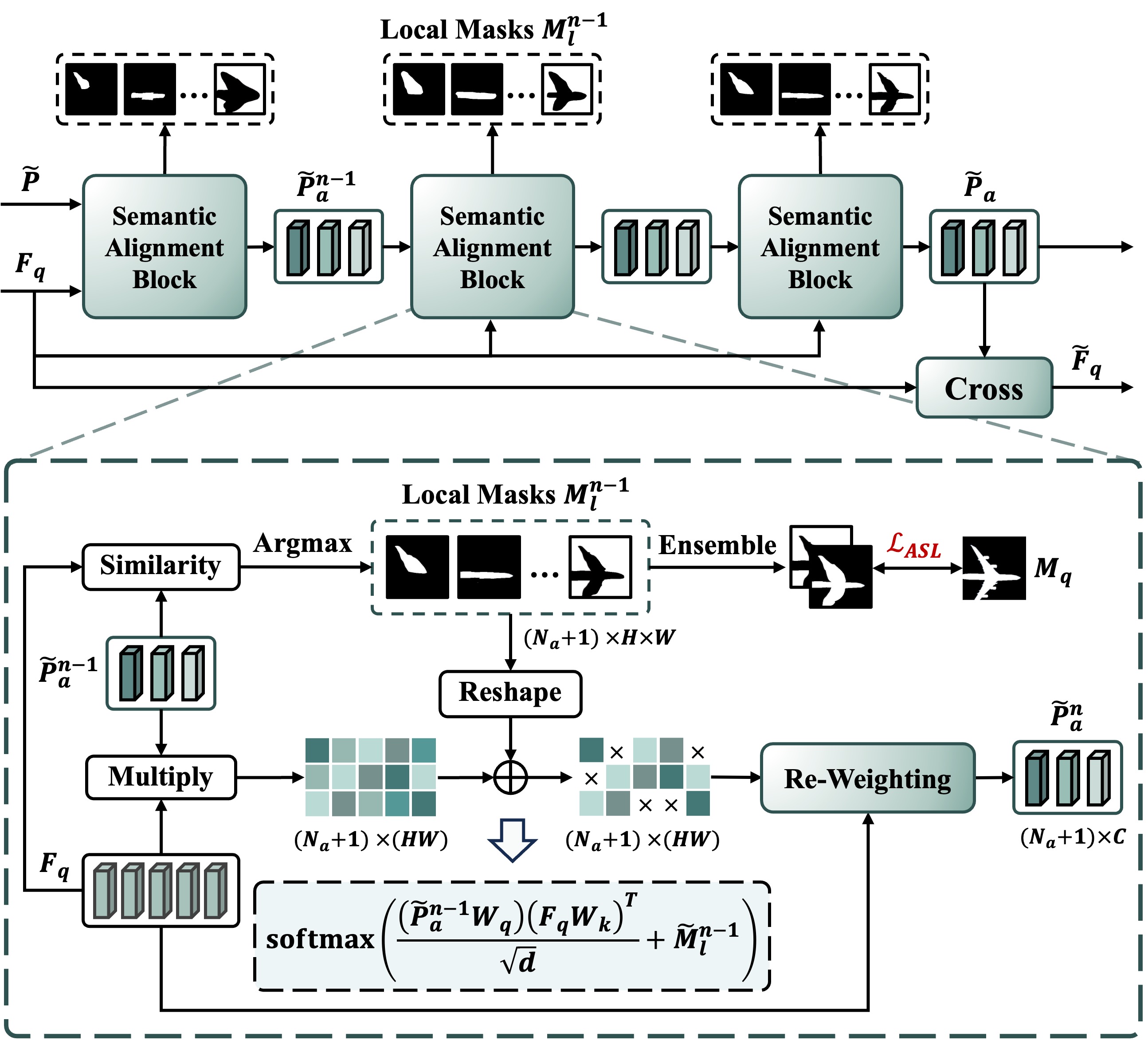}
\caption{\textbf{The detailed pipeline of Semantic Alignment Decoder (SAD),} which consists of multiple Semantic Alignment Blocks (SABs). For further aligning with the query object to bridge the semantic gap, SAB constructs the pseudo-local mask of the query image, which drives the agents to mine the local semantics of the query itself. Notably, We iterate the SAB process multiple times to gradually improve the quality of the pseudo-local mask, further promoting semantic consistency between the agents and the query object. $\widetilde{M}_l$ in the figure denotes the attention weight derived from the mapping the local mask $M_l$ through Eq.(\ref{equation:3}).}
\label{fig:5}
\end{figure*}

\normalem
\begin{algorithm}[t]
\caption{Semantic Alignment Decoder (SAD) }\label{Semantic Alignment Decoder}
\KwIn{ Local agents $\widetilde{P}$, Query feature $F_q$}
\KwOut{ Aligned agents $\widetilde{P}_a$, Aligned query feature $\widetilde{F}_q$}

\For{$i$ in $N$ }{
\textbf{Step 1 : Semantic Alignment Block}

Construct pseudo-local masks for the query image with local agents $M_l \leftarrow \left (\widetilde{P},F_q \right)$ according to Eq.(\ref{equation:10});

Different agents selectively aggregate query semantics under corresponding local masks $\widetilde{P}_a \leftarrow \left (\widetilde{P},F_q, M_l \right)$ according to Eq.(\ref{equation:11});
}

\textbf{Step 2:} The query feature aggregates the agents' semantics to enable semantic alignment $\widetilde{F}_q \leftarrow \left (F_q, \widetilde{P}_a \right)$.

\end{algorithm}
\ULforem

\subsection{Semantic Alignment Decoder}\label{sec:Semantic Alignment Decoder}
Based on the observation of strong intra-object similarity, i.e., pixels within the one object are more similar than pixels between different objects~\citep{wang2024focus}, we argue that it is also crucial to mine the valuable information from the query image itself, which has been omitted in previous work. Thus, the Semantic Alignment Decoder is proposed to drive the agents to mine the class-specific semantics from the query image for promoting semantic consistency with the query object, with the pipeline illustrated in Algorithm \ref{Semantic Alignment Decoder}.

\noindent \textbf{Semantic Alignment Block:} Concretely, as depicted in Fig.\ref{fig:5}, SAD consists of multiple Semantic Alignment Blocks (SABs) in series, each of which continuously aggregates the query semantics for agents via the cross-attention. We construct the pseudo-local masks of the query image to force different agents to selectively aggregate their desired local semantics without introducing irrelevant noise. More specifically, we perform similarity computation of query feature $F_q$ with different agents $\widetilde{p}^*$ to produce multiple segmentation results $y^*_l$, based on which we can gain the agent index corresponding to the maximum value of each position. Finally, the binary local masks $M_l \in \mathbb{R}{^{(N_a+1) \times H \times W}}$ corresponding to different agents can be derived by one-hot coding. The above process can be formulated as:
\begin{equation}\label{equation:10}
\setlength{\abovecaptionskip}{1pt}
\setlength{\belowcaptionskip}{1pt}
\begin{split}
&M_l = \mathrm{OneHot}\left ( \underset{i}{\operatorname{arg\,max}}\,  \left (\bigcup_{i=1}^{N_a+1} y^i_l \right )   \right )   \\
&s.t.\quad y^i_l=\left \langle F_q,\widetilde{p}^i \right \rangle 
\end{split}
\end{equation}
where $\mathrm{OneHot}\left(\cdot \right)$ denotes the process of encoding the local mask into binary masks, $\bigcup $ denotes the concatenation of multiple segmentation results, and $\mathrm{argmax}$ denotes the process of deriving the index corresponding to the maximum value at each pixel position. 

Under the constraint of these local masks $M_l = \left [ m_l^1,m_l^2,...,m_l^{(N_a+1)} \right ] $, different agents could selectively aggregate the local semantics within their respective corresponding masks in the query image without creating semantic confusion among agents. The above process can be formulated as:
\begin{align} \label{equation:11}
\setlength{\abovecaptionskip}{1pt}
\setlength{\belowcaptionskip}{1pt}
\widetilde{p}_a^i= \mathrm{softmax}\left ( \frac{\left ( \widetilde{p}^iW_q \right )\left ( F_qW_k \right )^T }{\sqrt{d} }+ \widetilde{m}_l^i  \right ) \left ( F_qW_v \right )   
\end{align}
where $\widetilde{P}_a=\left [\widetilde{p}_a^1,\widetilde{p}_a^2,...,\widetilde{p}_a^{\left (N_a+1  \right ) }\right ] $ denote the aligned agents, $\widetilde{p}^i$ denotes the $i^{th}$ agent, $F_q$ denotes the query feature, $\widetilde{m}_l^i$ denotes the attention weight derived from the mapping the $i^{th}$ local mask $m_l^i$ through Eq.(\ref{equation:3}), and $W_*$ denote the projection weights. Notably, the two-layer MLP in the semantic aggregation is omitted to simplify the formulation.

\noindent \textbf{Multiple block iterations:} Since the Semantic Alignment Block can drive agents to aggregate the corresponding local semantics in the query image to align with the query object. We can iterate the process multiple times to gradually improve the quality of the pseudo-local mask, thus further promoting semantic consistency between the agents and the query object. Suppose we have $N$-layer Semantic Alignment Blocks, then for each layer $n$:
\begin{align}  \label{equation:12}
\setlength{\abovecaptionskip}{1pt}
\setlength{\belowcaptionskip}{1pt}
\widetilde{P}_a^{n},M_l^{n-1}=\mathrm{SAB}\left ( \widetilde{P}_a^{n-1},F_q  \right ) , n=1,2,3,...,N
\end{align}
where $\mathrm{SAB}$ denotes the Semantic Alignment Block. Notably, after generating the last-layer aligned agents $\widetilde{P}_a$, we construct cross-attention to drive the query feature $\widetilde{F}_q$ to aggregate the agents' semantics, thus enabling semantic alignment.

\noindent \textbf{Agent Segmentation Loss:} To supervise that each Semantic Alignment Block works, we construct an Agent Segmentation Loss for constraint. Specifically, we ensemble the local segmentation results $\left \{y_l^i  \right \}_{i=1}^{N_a+1}$ received from each SAB to derive foreground-background binary prediction result, i.e., $\left [ y_l^{\left ( N_a+1 \right )}; \underset{i}{\operatorname{max}}\,  \left (\bigcup_{i=1}^{N_a} y^i_l \right )  \right ] $, where $y_l^{\left ( N_a+1 \right )}$ denotes the segmentation result from the background agent while other $y_l^i$ denote the segmentation result from other foreground agents. Here, we utilize binary cross-entropy (BCE) to constrain the prediction:
\begin{align}  \label{equation:13}
\setlength{\abovecaptionskip}{1pt}
\setlength{\belowcaptionskip}{1pt}
\mathcal{L}_{ASL} = \mathrm{BCE}\left (\left [ y_l^{\left ( N_a+1 \right )}; \underset{i}{\operatorname{max}}\,  \left (\bigcup_{i=1}^{N_a} y^i_l \right )  \right ],M_q \right )
\end{align}
Thus, with $N$-layer Semantic Alignment Blocks, $N$ losses are produced, where we take the mean as the final loss $\mathcal{L}_{ASL}$.

\subsection{Matching}\label{sec:Matching}

Given the final aligned agents $\widetilde{P}_a \in \mathbb{R}{^{(N_{a}+1) \times C}}$ and the aligned query feature $\widetilde{F}_q \in \mathbb{R}{^{H \times W \times C}}$, the segmentation result $\mathcal{P}\in \mathbb{R}{^{H\times W\times 2}}$ can be achieved by performing similarity computation between $\widetilde{F}_q$ and the agents $\widetilde{P}_a$ (including foreground agents $\left \{ \widetilde{p}_a^i \right \}_{i=1}^{N_a}$ and the background agent $\widetilde{p}_a^{N_{a}+1}$), which can be formulated as:
\begin{align}  \label{equation:14}
\setlength{\abovecaptionskip}{1pt}
\setlength{\belowcaptionskip}{1pt}
\mathcal{P}=\left [ \left \langle \widetilde{F}_q,\widetilde{P}_a^{N_a+1}  \right \rangle;  \underset{i}{\operatorname{max}}\,  \left (\bigcup_{i=1}^{N_a}\left \langle\widetilde{F}_q, \widetilde{P}_a^{i} \right \rangle  \right )
\right ] 
\end{align}
where $\left \langle \cdot,\cdot  \right \rangle $ denotes the similarity computation and $\bigcup $ denotes the concatenation of multiple segmentation results. Notably, the maximum operation $\mathrm{max} \left(\cdot \right)$ allows finding the foreground local agent that best matches the query pixel, thus effectively utilizing the local semantics to distinguish between the FG and BG pixels of the query image.

In this case, the binary cross entropy (BCE) loss $\mathcal{L}_{main}$ is employed to supervise the optimization of the segmentation prediction $\mathcal{P}$ towards $M_q$. Thus, the total loss $\mathcal{L}$ of the model consists of $\mathcal{L}_{main}$ and $\mathcal{L}_{ASL}$:
\begin{align}  \label{equation:15}
\setlength{\abovecaptionskip}{1pt}
\setlength{\belowcaptionskip}{1pt}
\mathcal{L}= \mathcal{L}_{main}+\gamma \mathcal{L}_{ASL}
\end{align}
where $\gamma$ aims to balance the contributions of $\mathcal{L}_{main}$ and $\mathcal{L}_{ASL}$.

\begin{figure*}[t]
\setlength{\abovecaptionskip}{1pt} \setlength{\belowcaptionskip}{1pt}
\centering
\includegraphics[width=1.0\linewidth]{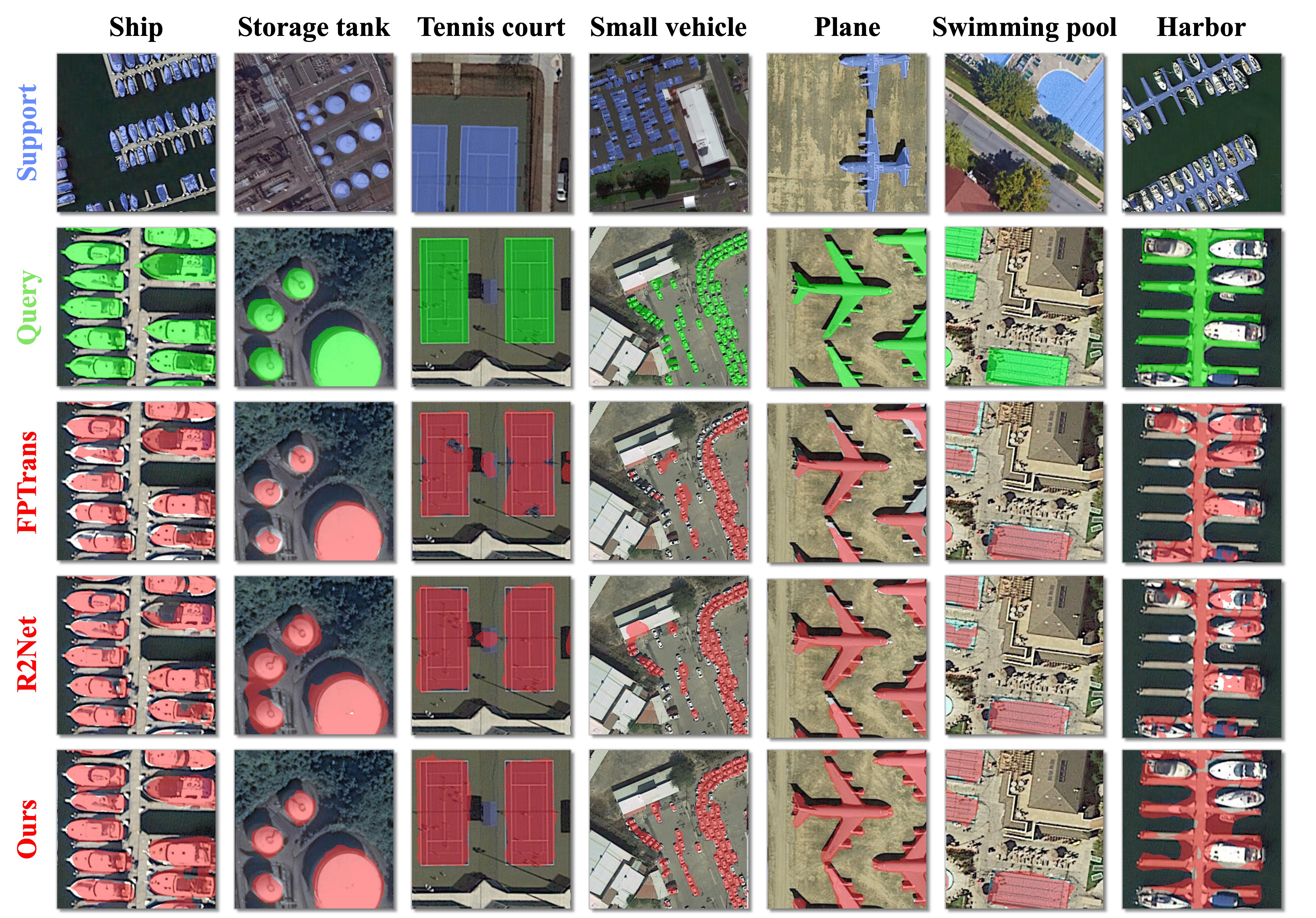}
\caption{\textbf{Qualitative visualization of our AgMTR and several competing methods under the 1-shot setting on iSAID benchmark.} From top to bottom, each row represents: support images with GT (\textcolor{blue}{blue}), query images with GT (\textcolor{green}{green}), segmentation results of FPTrans (\textcolor{red}{red}), segmentation results of R2Net (\textcolor{red}{red}), and our segmentation results (\textcolor{red}{red}).}
\label{fig:6}
\end{figure*}

\begin{table*}[t]
\setlength{\abovecaptionskip}{1pt}
\setlength{\belowcaptionskip}{1pt}
\caption{ Performance comparison on iSAID benchmark measured in mIoU and FB-IoU($\%$) under the 1- and 5-shot settings. The `Mean' and `FB-IoU' indicate the averaged mIoU accuracy and the averaged FB-IoU accuracy for all classes under all folds, respectively. } \label{tab:1}
\renewcommand\arraystretch{1.5}
\centering
\resizebox{1.0\linewidth}{!}{
\begin{tabular}{cr|ccc|cc|ccc|cc} 
\toprule
\multirow{2}{*}{Backbone}    & \multirow{2}{*}{Method}                   & \multicolumn{5}{c|}{1-shot}   & \multicolumn{5}{c}{5-shot}      \\ 
\cline{3-12}
&                                                  & Fold-0                                             & Fold-1                                             & Fold-2                                             & Mean $\uparrow$                                              & FB-IoU $\uparrow$        & Fold-0                                             & Fold-1                                             & Fold-2                                             & Mean $\uparrow$                                              & FB-IoU $\uparrow$         \\ 
\hline
\multirow{11}{*}{ResNet-50}  & SCL (CVPR2021)~\citep{zhang2021self}                                 & 49.08                                              & 35.61                                              & 48.15                                              & 44.28                                              & 61.66          & 50.69                                              & 35.64                                              & 48.72                                              & 45.02                                              & 62.63           \\
& ASGNet (CVPR2021)~\citep{li2021adaptive}                              & 48.59                                              & 36.82                                              & 46.65                                              & 44.02                                              & 62.36          & 53.01                                              & 37.44                                              & 52.18                                              & 47.54                                              & 64.63           \\
& CyCTR (NeurIPS2021)~\citep{zhang2021few}                            & 51.15                                              & 38.40                                              & 53.79                                              & 47.78                                              & 62.86          & 51.91                                              & 39.01                                              & 54.83                                              & 48.58                                              & 63.81           \\

& PFENet (TPAMI2022)~\citep{tian2022prior}                              & 51.34                                              & 38.79                                              & 52.26                                              & 47.46                                              & 63.34          & 54.71                                              & 41.51                                              & 54.45                                              & 50.22                                              & 64.99           \\
& NTRENet (CVPR2022) ~\citep{liu2022learning}                              & 49.52                                              & 38.66                                              & 51.87                                              & 46.68                                              & 62.84          & 50.60                                              & 40.99                                              & 55.07                                              & 48.89                                              & 63.74           \\
& DCPNet (IJCV2023)~\citep{lang2024few}                               & 48.43                                              & 37.59                                              & 52.09                                              & 46.04                                              & 62.76          & 50.34                                              & 40.75                                              & 51.33                                              & 47.47                                              & 64.10           \\
& SCCAN (ICCV2023)~\citep{xu2023self}                               & 52.24                                              & 38.60                                              & 52.84                                              & 47.89                                              & 63.26          & 52.75                                              & 38.95                                              & 54.26                                              & 48.65                                              & 62.23           \\
& BAM (TPAMI2023)~\citep{lang2023base}    & 58.17  & 42.12 &50.95  &50.41 &67.03              &62.51                                                  &43.16  &58.24                                                  &54.64                                                &67.92               \\
& R2Net (TGRS2023)~\citep{lang2023global}   & 56.81  & 39.85 &49.02  &48.56 &62.65              &60.47                                                  &41.43  &50.24                                                  &50.71                                                  &65.70               \\
& DMNet (TGRS2023) ~\citep{bi2023not}                               & 54.45                                              & 40.68                                              & 53.60                                              & 49.58                                              & 64.46          & 57.67                                              & 41.06                                              & 55.28                                              & 51.34                                              & 65.81           \\ 
\hline
\multirow{10}{*}{ResNet-101} & SCL (CVPR2021)~\citep{zhang2021self}                                   & 47.59                                              & 36.90                                              & 45.21                                              & 43.23                                              & 61.71          & 48.94                                              & 38.01                                              & 46.21                                              & 44.39                                              & 60.93           \\
& ASGNet (CVPR2021) ~\citep{li2021adaptive}                              & 47.55                                              & 38.47                                              & 49.28                                              & 45.10                                              & 62.03          & 53.54                                              & 38.24                                              & 53.20                                              & 48.33                                              & 65.35           \\
& CyCTR (NeurIPS2021)~\citep{zhang2021few}                            & 50.89                                              & 38.89                                              & 52.22                                              & 47.33                                              & 62.35          & 52.15                                              & 40.28                                              & 55.32                                              & 49.25                                              & 64.45           \\
& PFENet (TPAMI2022)~\citep{tian2022prior}                             & 50.69                                              & 38.37                                              & 52.85                                              & 47.30                                              & 62.46          & 54.40                                              & 41.55                                              & 50.55                                              & 48.83                                              & 64.57           \\
& NTRENet (CVPR2022) ~\citep{liu2022learning}                             & 50.33                                              & 38.73                                              & 51.23                                              & 46.76                                              & 63.25          & 53.24                                              & 41.87                                              & 51.53                                              & 48.88                                              & 64.16           \\
& DCPNet (IJCV2023)~\citep{lang2024few}                              & 47.63                                              & 38.80                                              & 49.34                                              & 45.26                                              & 62.56          & 50.68                                              & 40.02                                              & 54.52                                              & 48.41                                              & 62.96           \\
& SCCAN (ICCV2023)~\citep{xu2023self}                               & 52.71                                              & 38.76                                              & 53.45                                              & 48.31                                              & 62.72          & 53.85                                              & 39.11                                              & 54.56                                              & 49.17                                              & 64.27           \\
& BAM (TPAMI2023)~\citep{lang2023base}                                 & 59.10  & 41.65 &50.54  &50.43 &66.05              &61.57                                                  &42.32  &55.34                                                  &53.08                                                  &66.42               \\
& R2Net (TGRS2023)~\citep{lang2023global}                               & 57.74  & 39.76 &48.45  &48.65 &63.38              &58.83                                                  &40.58  &49.69                                                  &49.70                                                  &63.79           \\
& DMNet (TGRS2023)~\citep{bi2023not}                               & 54.01                                              & 40.04                                              & 53.57                                              & 49.21                                              & 64.03          & 55.70                                              & 41.69                                              & 56.47                                              & 51.29                                              & 65.88           \\ 
\hline
\multirow{2}{*}{ViT-B/16}    & FPTrans (NeurIPS2022)~\citep{zhang2022feature}                          & 50.04                                              & 36.85                                              & 53.30                                              & 46.73                                              & 62.36          & 57.92                                              & 47.42                                              & 64.23                                              & 56.52                                              & 66.12           \\
& {\cellcolor[rgb]{0.925,0.925,0.925}}\textbf{AgMTR (Ours)} & {\cellcolor[rgb]{0.925,0.925,0.925}}\textbf{55.45} & {\cellcolor[rgb]{0.925,0.925,0.925}}\textbf{40.67} & {\cellcolor[rgb]{0.925,0.925,0.925}}\textbf{53.81} & {\cellcolor[rgb]{0.925,0.925,0.925}}\textbf{49.98} & {\cellcolor[rgb]
{0.925,0.925,0.925}}\textbf{63.20}  & {\cellcolor[rgb]{0.925,0.925,0.925}}\textbf{59.05} & {\cellcolor[rgb]{0.925,0.925,0.925}}\textbf{47.86} & {\cellcolor[rgb]{0.925,0.925,0.925}}\textbf{64.08} & {\cellcolor[rgb]{0.925,0.925,0.925}}\textbf{57.00} &{\cellcolor[rgb]{0.925,0.925,0.925}} \textbf{67.60}   \\ 
\hline
\multirow{2}{*}{DeiT-B/16}   & FPTrans (NeurIPS2022)~\citep{zhang2022feature}                          & 51.59                                              & 37.79                                              & 52.85                                              & 47.41                                              & 62.52          & 58.52                                              & 47.24                                              & 65.37                                              & 57.04                                              & 67.63           \\
& {\cellcolor[rgb]{0.925,0.925,0.925}}\textbf{AgMTR (Ours)} & {\cellcolor[rgb]{0.925,0.925,0.925}}\textbf{57.51} & {\cellcolor[rgb]{0.925,0.925,0.925}}\textbf{43.54} & {\cellcolor[rgb]{0.925,0.925,0.925}}\textbf{53.69} & {\cellcolor[rgb]{0.925,0.925,0.925}}\textbf{51.58} & {\cellcolor[rgb]{0.925,0.925,0.925}}\textbf{64.63} & {\cellcolor[rgb]{0.925,0.925,0.925}}\textbf{61.51} & {\cellcolor[rgb]{0.925,0.925,0.925}}\textbf{49.23} & {\cellcolor[rgb]{0.925,0.925,0.925}}\textbf{64.05} & {\cellcolor[rgb]{0.925,0.925,0.925}}\textbf{58.26} &{\cellcolor[rgb]{0.925,0.925,0.925}} \textbf{68.72}  \\
\bottomrule
\end{tabular}}
\end{table*}

\section{Experiments}\label{sec:Experiments}

Firstly, Sec.\ref{sec: Experimental Setup} describes the experimental setup, including dataset introduction, implementation details, and evaluation metrics. Then, Sec.\ref{sec: Comparison with State-of-the-arts} and Sec.\ref{sec: Ablation Study and Analysis} conduct extensive comparative experiments and ablation studies to prove the superiority of our method, respectively. Finally, Sec.\ref{sec: Failure case analysis} provides several failure cases.

\subsection{Experimental Setup}\label{sec: Experimental Setup}
\subsubsection{Datasets}\label{sec:Datasets} 

Three FSS benchmarks are utilized for evaluation, including a remote sensing benchmark iSAID~\citep{bi2023not} and two CV benchmarks PASCAL-$5^i$~\citep{10256677} and COCO-$20^i$~\citep{tian2022prior}.


\noindent \textbf{iSAID}~\citep{waqas2019isaid}, a massive benchmark for intensive prediction tasks in remote sensing scenarios, contains 655,451 objects from 15 classes in 2,806 high-resolution images. Following the setup of DMNet~\citep{bi2023not}, this dataset can be utilized for evaluating the FSS task. Concretely, the 15 classes will be split into three folds, each containing 10 training classes and 5 testing classes. Remarkably, both training and testing classes are disjointed in each fold to build unseen class situations. For each fold, we randomly sample 1000 image pairs for validation. 

\noindent \textbf{PASCAL-$5^i$}, which is constructed from PASCAL VOC 2012~\citep{everingham2010pascal} and additional SBD~\citep{hariharan2014simultaneous}, contains 20 classes in total. For cross-validation, these classes will be split into four folds, each containing 15 training classes and 5 testing classes as done in OSLSM~\citep{shaban2017one}. For each fold, we randomly sample 5000 image pairs for validation. 

\noindent \textbf{COCO-$20^i$}, which is constructed from MS COCO~\citep{lin2014microsoft}, contains 80 classes in total. For cross-validation, these classes will be split into four folds, each containing 60 training classes and 20 testing classes as done in~\citep{lang2024few}. For each fold, we randomly sample 5000 image pairs for validation. 

\subsubsection{Implementation Details}\label{sec:Implementation Details}

The model\footnote[1]{The codes are available at \href{https://github.com/HanboBizl/AgMTR/}{https://github.com/HanboBizl/AgMTR/}.} is implemented on Tesla A40 GPUs using the PyTorch framework~\citep{paszke2019pytorch}. The SGD optimizer is utilized to optimize the model during the training phase, with the weight decay set to 0.00005. The batch size is set to 8 and the initial learning rate is set to 0.002, where the poly strategy~\citep{chen2017deeplab} is utilized to adjust the learning rate. For iSAID, we resize the images to 256$\times$256 and set 40 epochs. For PASCAL-$5^i$, we resize the images to 480$\times$480 and set 30 epochs, and for COCO-$20^i$, we resize the images to 480$\times$480 and set 40 epochs. The same data enhancements are utilized to enable fair comparisons with previous work.

For the proposed model, two transformers, i.e., ViT~\citep{dosovitskiy2020image} and DeiT~\citep{touvron2021training}, are utilized as the feature encoder, and both of them are pre-trained on ImageNet~\citep{deng2009imagenet}. In the experiment, the number of agents $N_a$, unlabeled images $N_u$, pixel sets obtained through SLIC $N_s$, and SAB layers $N$ are set to 5, 5, 100, and 7, respectively. The weight $\gamma$ to balance two losses is set to 0.8.

\subsubsection{Evaluation Metrics}\label{sec:Evaluation Metrics}
Following previous FSS schemes~\citep{lang2023base}, we adopt the class mean intersection-over-union (mIoU) and foreground-background IoU (FB-IoU) as the evaluation metrics. The former calculates the averaged segmentation accuracy between different classes, while the latter calculates the accuracy between foreground and background. The mIoU evaluation metric is calculated as: $\mathrm{mIoU}=\frac{1}{n}\sum_{i=1}^{N_\mathrm{unseen}}\mathrm{IoU_i}$, where $N_\mathrm{unseen}$ indicates the number of unseen classes in each fold and $\mathrm{IoU_i}$ indicates the IoU score of the class $i$. And the FB-IoU metric is calculated as: $\mathrm{FB}$-$\mathrm{IoU} =\left ( \mathrm{IoU_{F}}+ \mathrm{IoU_{B}}  \right )/2$, where $\mathrm{IoU_F}$ and $\mathrm{IoU_B}$ indicate the foreground and background IoU scores, respectively. 

\begin{table*}[t]
\setlength{\abovecaptionskip}{1pt}
\setlength{\belowcaptionskip}{1pt}
\caption{ Performance comparison on PASCAL-$5^i$ benchmark measured in mIoU ($\%$) under the 1- and 5-shot settings. The `Mean' indicates the averaged mIoU accuracy for all classes under all folds.} \label{tab:2}
\renewcommand\arraystretch{1.5}
\centering
\resizebox{1.0\linewidth}{!}{
\begin{tabular}{cr|cccc|c|cccc|c} 
\toprule
\multirow{2}{*}{Backbone}   & \multirow{2}{*}{Method} & \multicolumn{5}{c|}{1-shot}                                                        & \multicolumn{5}{c}{5-shot}                                                          \\ 
\cline{3-12}
&                         & Fold-0         & Fold-1         & Fold-2         & Fold-3         & Mean $\uparrow$           & Fold-0         & Fold-1         & Fold-2         & Fold-3         & Mean $\uparrow$           \\ 
\hline
\multirow{12}{*}{ResNet-50} & CyCTR (NeurIPS2021)~\citep{zhang2021few}    & 65.70          & 71.00          & 59.50          & 59.70          & 63.98          & 69.30          & 73.50          & 63.80          & 63.50          & 67.53           \\
& PFENet (TPAMI2022)~\citep{tian2022prior}    & 61.70          & 69.50          & 55.40          & 56.30          & 60.73          & 63.10          & 70.70          & 55.80          & 57.90          & 61.88           \\
& HPA (TPAMI2022)~\citep{cheng2022holistic}       & 65.94          & 71.96          & 64.66          & 56.78          & 64.84          & 70.54          & 73.28          & 68.37          & 63.41          & 68.90           \\
& AAFormer (ECCV2022)~\citep{wang2022adaptive}   & 69.10          & 73.30          & 59.10          & 59.20          & 65.18          & 72.50          & 74.70          & 62.00          & 61.30          & 67.63           \\
& NTRENet (CVPR2022)~\citep{liu2022learning}    & 65.40          & 72.30          & 59.40          & 59.80          & 64.23          & 66.20          & 72.80          & 61.70          & 62.20          & 65.73           \\
& IPMT (NeurIPS2022)~\citep{liu2022intermediate}    & 72.80          & 73.70          & 59.20          & 61.60          & 66.83          & 73.10          & 74.70          & 61.60          & 63.40          & 68.20           \\ 
& BAM (TPAMI2023)~\citep{lang2023base}       & 69.15          & 74.66          & 67.79          & 61.72          & 68.33          & 71.84          & 75.67          & 71.95          & 67.54          & 71.75           \\
& DCPNet (IJCV2023)~\citep{lang2024few}     & 67.20          & 72.90          & 65.20          & 59.40          & 66.10          & 70.50          & 75.30          & 68.00          & 67.70          & 70.30           \\
& SCCAN (ICCV2023)~\citep{xu2023self}      & 68.30          & 72.50          & 66.80          & 59.80          & 66.80          & 72.30          & 74.10          & 69.10          & 65.60          & 70.30           \\
& HDMNet (CVPR2023)~\citep{peng2023hierarchical}     & 71.00          & 75.40          & 68.90          & 62.10          & 69.35          & 71.30          & 76.20          & 71.30          & 68.50          & 71.83           \\
& MIANet (CVPR2023)~\citep{yang2023mianet}     & 68.51          & 75.76          & 67.46          & 63.15          & 68.72          & 70.20          & 77.38          & 70.02          & 68.77          & 71.59           \\
& RiFeNet (AAAI2024)~\citep{bao2023relevant}    & 68.40          & 73.50          & 67.10          & 59.40          & 67.10          & 70.00          & 74.70          & 69.40          & 64.20          & 69.60           \\ 
\hline
\multirow{9}{*}{ResNet-101} & CyCTR (NeurIPS2021)~\citep{zhang2021few}   & 67.20          & 71.10          & 57.60          & 59.00          & 63.73          & 71.00          & 75.00          & 58.50          & 65.00          & 67.38           \\
& PFENet (TPAMI2022)~\citep{tian2022prior}    & 60.50          & 69.40          & 54.40          & 55.90          & 60.05          & 62.80          & 70.40          & 54.90          & 57.60          & 61.43           \\
& AAFormer (ECCV2022)~\citep{wang2022adaptive}    & 69.90          & 73.60          & 57.90          & 59.70          & 65.28          & 75.00          & 75.10          & 59.00          & 63.20          & 68.08           \\
& NTRENet (CVPR2022)~\citep{liu2022learning}    & 65.50          & 71.80          & 59.10          & 58.30          & 63.68          & 67.90          & 73.20          & 60.10          & 66.80          & 67.00           \\
& IPMT (NeurIPS2022)~\citep{liu2022intermediate}    & 71.60          & 73.50          & 58.00          & 61.20          & 66.08          & 75.30          & 76.90          & 59.60          & 65.10          & 69.23           \\
& HPA (TPAMI2022)~\citep{cheng2022holistic}       & 66.40          & 72.65          & 64.10          & 59.42          & 65.64          & 68.02          & 74.60          & 65.85          & 67.11          & 68.90           \\
& BAM (TPAMI2023)~\citep{lang2023base}       & 69.88          & 75.37          & 67.06          & 62.11          & 68.61          & 72.62          & 77.06          & 70.67          & 69.82          & 72.54           \\
& DCPNet (IJCV2023)~\citep{lang2024few}      & 68.90          & 74.20          & 63.30          & 62.70          & 67.30          & 72.10          & 77.10          & 66.50          & 70.50          & 71.50           \\
& SCCAN (ICCV2023)~\citep{xu2023self}      & 70.90          & 73.90          & 66.80          & 61.70          & 68.30          & 73.10          & 76.40          & 70.30          & 66.10          & 71.50           \\ 
\hline
\multirow{2}{*}{ViT-B/16}   & FPTrans (NeurIPS2022)~\citep{zhang2022feature} & 67.10          & 69.80          & 65.60          & 56.40          & 64.73          & 73.50          & 75.70          & 77.40          & 68.30          & 73.73           \\
& {\cellcolor[rgb]{0.925,0.925,0.925}}\textbf{AgMTR (Ours)}           & {\cellcolor[rgb]{0.925,0.925,0.925}}\textbf{70.17} & {\cellcolor[rgb]{0.925,0.925,0.925}}\textbf{74.68} & {\cellcolor[rgb]{0.925,0.925,0.925}}\textbf{67.15} & {\cellcolor[rgb]{0.925,0.925,0.925}}\textbf{58.20} & {\cellcolor[rgb]{0.925,0.925,0.925}}\textbf{67.55} & {\cellcolor[rgb]{0.925,0.925,0.925}}\textbf{73.55} & {\cellcolor[rgb]{0.925,0.925,0.925}}\textbf{76.84} & {\cellcolor[rgb]{0.925,0.925,0.925}}\textbf{74.75} & {\cellcolor[rgb]{0.925,0.925,0.925}}\textbf{66.78} & {\cellcolor[rgb]{0.925,0.925,0.925}}\textbf{72.98}  \\ 
\hline
\multirow{3}{*}{DeiT-B/16}  & FPTrans (NeurIPS2022)~\citep{zhang2022feature} & 72.30          & 70.60          & 68.30          & 64.10          & 68.83          & 76.70          & 79.00          & 81.00          & 75.10          & 77.95           \\
& MuHS (ICLR2023)~\citep{hu2022suppressing}       & 71.20          & 71.50          & 67.00          & 66.60          & 69.08          & 75.70          & 77.80          & 78.60          & 74.70          & 76.70           \\
& {\cellcolor[rgb]{0.925,0.925,0.925}}\textbf{AgMTR (Ours)}           & {\cellcolor[rgb]{0.925,0.925,0.925}}\textbf{72.13} & {\cellcolor[rgb]{0.925,0.925,0.925}}\textbf{75.72} & {\cellcolor[rgb]{0.925,0.925,0.925}}\textbf{66.31} & {\cellcolor[rgb]{0.925,0.925,0.925}}\textbf{66.91} & {\cellcolor[rgb]{0.925,0.925,0.925}}\textbf{70.27} & {\cellcolor[rgb]{0.925,0.925,0.925}}\textbf{76.44} & {\cellcolor[rgb]{0.925,0.925,0.925}}\textbf{78.54} & {\cellcolor[rgb]{0.925,0.925,0.925}}\textbf{75.56} & {\cellcolor[rgb]{0.925,0.925,0.925}}\textbf{74.95} & {\cellcolor[rgb]{0.925,0.925,0.925}}\textbf{76.37}  \\
\bottomrule
\end{tabular}}
\end{table*}

\begin{table*}[t]
\setlength{\abovecaptionskip}{1pt}
\setlength{\belowcaptionskip}{1pt}
\caption{ Performance comparison on COCO-$20^i$ benchmark measured in mIoU ($\%$) under the 1- and 5-shot settings. The `Mean' indicates the averaged mIoU accuracy for all classes under all folds.} \label{tab:3}
\renewcommand\arraystretch{1.5}
\centering
\resizebox{1.0\linewidth}{!}{
\begin{tabular}{cr|cccc|c|cccc|c} 
\toprule
\multirow{2}{*}{Backbone}   & \multirow{2}{*}{Method} & \multicolumn{5}{c|}{1-shot}                                                        & \multicolumn{5}{c}{5-shot}                                                          \\ 
\cline{3-12}
&                         & Fold-0         & Fold-1         & Fold-2         & Fold-3         & Mean $\uparrow$          & Fold-0         & Fold-1         & Fold-2         & Fold-3         & Mean $\uparrow$           \\ 
\hline
\multirow{11}{*}{ResNet-50} & CyCTR (NeurIPS2021)~\citep{zhang2021few}    & 38.90          & 43.00          & 39.60          & 39.80          & 40.33          & 41.10          & 48.90          & 45.20          & 47.00          & 45.55           \\
& PFENet (TPAMI2022)~\citep{tian2022prior}     & 36.50          & 38.60          & 34.50          & 33.80          & 35.85          & 36.50          & 43.30          & 37.80          & 38.40          & 39.00           \\
& HPA (TPAMI2022)~\citep{cheng2022holistic}       & 40.30          & 46.57          & 44.12          & 42.71          & 43.43          & 45.54          & 55.43          & 48.90          & 50.21          & 50.02           \\
& AAFormer (ECCV2022)~\citep{wang2022adaptive}   & 39.80          & 44.60          & 40.60          & 41.40          & 41.60          & 42.90          & 50.10          & 45.50          & 49.20          & 46.93           \\
& NTRENet (CVPR2022)~\citep{liu2022learning}    & 36.80          & 42.60          & 39.90          & 37.90          & 39.30          & 38.20          & 44.10          & 40.40          & 38.40          & 40.28           \\
& IPMT (NeurIPS2022)~\citep{liu2022intermediate}     & 41.40          & 45.10          & 45.60          & 40.00          & 43.03          & 43.50          & 49.70          & 48.70          & 47.90          & 47.45           \\
& BAM (TPAMI2023)~\citep{lang2023base}       & 43.86          & 51.38          & 47.86          & 44.47          & 46.89          & 49.76          & 55.43          & 52.31          & 50.19          & 51.92           \\
& DCPNet (IJCV2023)~\citep{lang2024few}      & 43.00          & 48.60          & 45.40          & 44.80          & 45.50          & 47.00          & 54.70          & 51.70          & 50.00          & 50.90           \\
& SCCAN (ICCV2023)~\citep{xu2023self}      & 40.40          & 49.70          & 49.60          & 45.60          & 46.30          & 47.20          & 57.20          & 59.20          & 52.10          & 53.90           \\
& MIANet (CVPR2023)~\citep{yang2023mianet}     & 42.49          & 52.95          & 47.77          & 47.42          & 47.66          & 45.84          & 58.18          & 51.29          & 51.90          & 51.80           \\
& RiFeNet (AAAI2024)~\citep{bao2023relevant}    & 39.10          & 47.20          & 44.60          & 45.40          & 44.10          & 44.30          & 52.40          & 49.30          & 48.40          & 48.60           \\ 
\hline
\multirow{5}{*}{ResNet-101} & PFENet (TPAMI2022)~\citep{tian2022prior}    & 34.40          & 33.00          & 32.30          & 30.10          & 32.45          & 38.50          & 38.60          & 38.20          & 34.30          & 37.40           \\
& NTRENet (CVPR2022)~\citep{liu2022learning}    & 38.30          & 40.40          & 39.50          & 38.10          & 39.08          & 42.30          & 44.40          & 44.20          & 41.70          & 43.15           \\
& IPMT (NeurIPS2022)~\citep{liu2022intermediate}    & 40.50          & 45.70          & 44.80          & 39.30          & 42.58          & 45.10          & 50.30          & 49.30          & 46.80          & 47.88           \\
& HPA (TPAMI2022)~\citep{cheng2022holistic}       & 43.08          & 50.01          & 44.78          & 45.19          & 45.77          & 49.18          & 57.76          & 51.97          & 50.64          & 52.39           \\
& SCCAN (ICCV2023)~\citep{xu2023self}      & 42.60          & 51.40          & 50.00          & 48.80          & 48.20          & 49.40          & 61.70          & 61.90          & 55.00          & 57.00           \\ 
\hline
\multirow{3}{*}{DeiT-B/16}  & FPTrans (NeurIPS2022)~\citep{zhang2022feature}  & 44.40          & 48.90          & 50.60          & 44.00          & 46.98          & 54.20          & 62.50          & 61.30          & 57.60          & 58.90           \\
& MuHS (ICLR2023)~\citep{hu2022suppressing}       & 44.00          & 50.00          & 49.10          & 46.30          & 47.35          & 53.60          & 60.50          & 57.30          & 55.20          & 56.65           \\
& {\cellcolor[rgb]{0.925,0.925,0.925}}\textbf{AgMTR (Ours)}           & {\cellcolor[rgb]{0.925,0.925,0.925}}\textbf{44.01} & {\cellcolor[rgb]{0.925,0.925,0.925}}\textbf{55.51} & {\cellcolor[rgb]{0.925,0.925,0.925}}\textbf{49.90} & {\cellcolor[rgb]{0.925,0.925,0.925}}\textbf{46.61} & {\cellcolor[rgb]{0.925,0.925,0.925}}\textbf{49.01} & {\cellcolor[rgb]{0.925,0.925,0.925}}\textbf{52.51} & {\cellcolor[rgb]{0.925,0.925,0.925}}\textbf{62.10} & {\cellcolor[rgb]{0.925,0.925,0.925}}\textbf{59.64} & {\cellcolor[rgb]{0.925,0.925,0.925}}\textbf{53.61} & {\cellcolor[rgb]{0.925,0.925,0.925}}\textbf{56.97}  \\
\bottomrule
\end{tabular}}
\end{table*}

\begin{figure}[t]
\setlength{\abovecaptionskip}{1pt} \setlength{\belowcaptionskip}{1pt}
\centering
\includegraphics[width=1.0\linewidth]{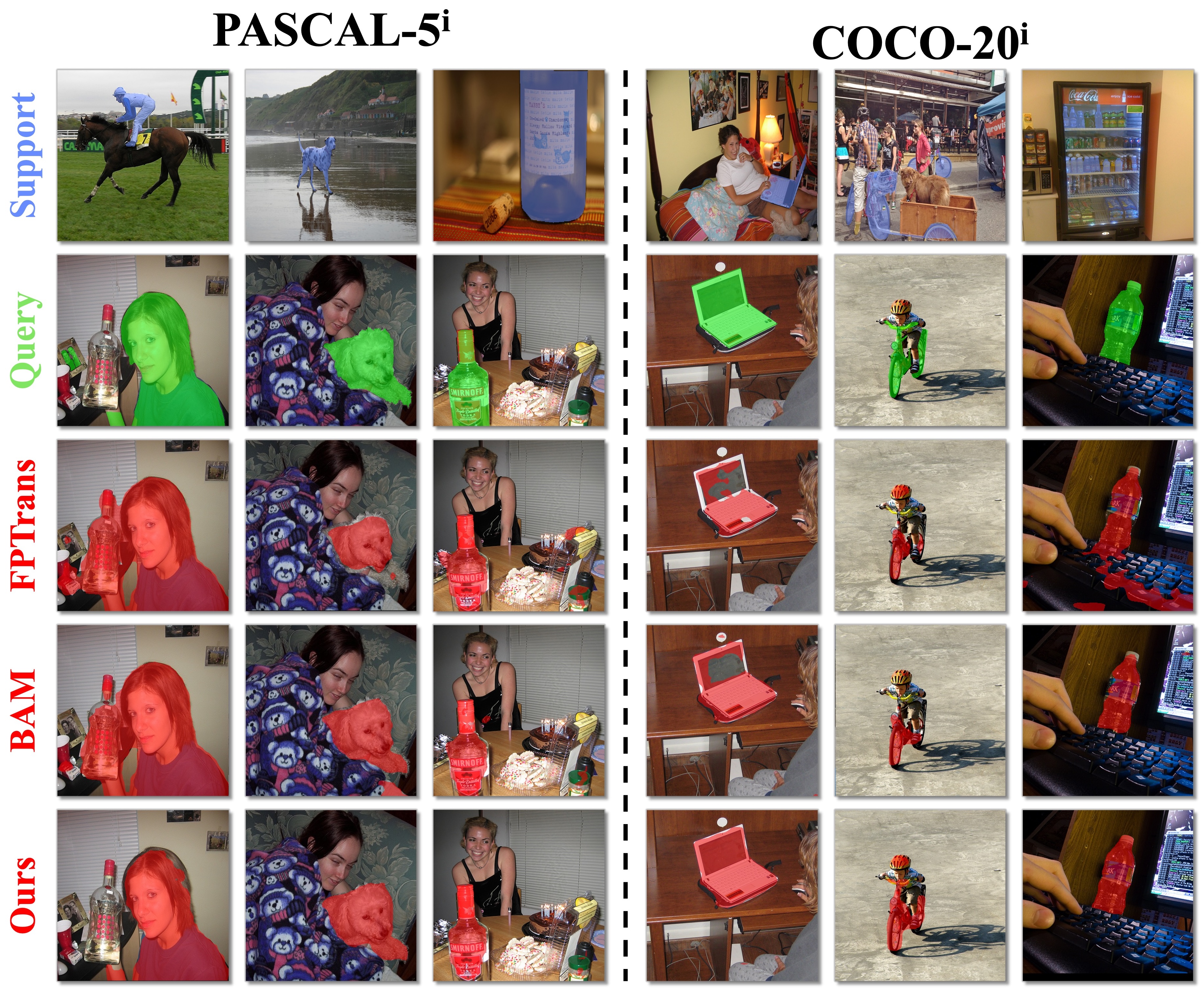}
\caption{\textbf{Qualitative visualization of our AgMTR and several competing methods under the 1-shot setting on more common natural scenarios.} The left panel shows the PASCAL-$5^i$ results while the right shows the COCO-$20^i$ results. From top to bottom, each row represents: support images with GT (\textcolor{blue}{blue}), query images with GT (\textcolor{green}{green}), segmentation results of FPTrans (\textcolor{red}{red}), segmentation results of BAM (\textcolor{red}{red}), and our segmentation results (\textcolor{red}{red}).}
\label{fig:7}
\end{figure}

\subsection{Comparison with State-of-the-arts}\label{sec: Comparison with State-of-the-arts}

\subsubsection{Quantitative Results}\label{sec:Quantitative Results}
\textbf{iSAID: } Table \ref{tab:1} quantitatively depicts the comparison between the proposed and other methods on iSAID benchmark. Satisfyingly, our AgMTR achieves state-of-the-art performance in all cases. For example, under the 1-shot setting, AgMTR with DeiT-B/16 backbone achieves 51.58\% mIoU, outperforming the best CNN-based method BAM~\citep{lang2023base} by 1.15\% and the Transformer-based method FPTrans~\citep{zhang2022feature} by 4.17\%. Surprisingly, AgMTR under the 1-shot setting even beats several methods under the 5-shot setting, e.g., R2Net~\citep{lang2023global}, DMNet~\citep{bi2023not}, and SCCAN~\citep{xu2023self}, which fully demonstrates that our method can alleviate the pressure of dense annotation and easily generalize to unseen classes. Under the 5-shot setting, AgMTR achieves 3.62\% gains over the best BAM. Similar gains are realized in FB-IoU. The above results strongly prove that our agent-level semantic correlation can effectively handle large intra-class variations and complex backgrounds in remote sensing scenarios, realizing excellent segmentation performance for unseen classes.

\noindent \textbf{PASCAL-$5^i$ and COCO-$20^i$: } We also explore the performance of the proposed method in more common natural scenarios to further explore its generalization and applicability. Table \ref{tab:2}-\ref{tab:3} present the 1- and 5-shot results on PASCAL-$5^i$ and COCO-$20^i$. Encouragingly, our AgMTR performs equally competitively. Specifically, for PASCAL-$5^i$, AgMTR with DeiT-16/B backbone outperforms the top CNN method HDMNet~\citep{peng2023hierarchical} by 0.92\% and 4.54\% under the 1- and 5-shot settings, respectively. Besides, in comparison with the transformer-based methods FPTrans~\citep{zhang2022feature} and MuHS~\citep{hu2022suppressing}, we realize a lead of 1.44\% and 1.19\% mIoU under the 1-shot setting. And for COCO-$20^i$ with larger object differences and more complex backgrounds, similar gains can be realized as well. For instance, AgMTR exceeds the top CNN-based method MIANet~\citep{yang2023mianet} and the Transformer-based method MuHS~\citep{hu2022suppressing} by 0.81\% and 1.66\%, respectively, under the 1-shot setting. The above results indicate that our method is general enough to be applied to various domains.

\subsubsection{Qualitative Results}\label{sec:Qualitative Results}

To further exhibit the excellence of the proposed AgMTR, we visualize several segmentation results on three FSS datasets. Fig.\ref{fig:6} illustrates the qualitative results on iSAID. One can find that our method has clear advantages: (i) Benefiting from mining class-specific semantics from support images, unlabeled images, and the query image, our agents have more powerful class-awareness to perform precise agent-level semantic correlation, thus effectively focusing on the interested class while suppressing the activation of other classes. For instance, for the dense small targets `small vehicles' in $4^{th}$ column and the `harbor' with irrelevant classes interference in $7^{th}$ column, R2Net incorrectly activates the `Roof' and `Boats', respectively, while our AgMTR segments nicely. (ii) Benefiting from the local semantic exploration of the target category, our AgMTR can reasonably find the local agent that best matches the query pixel, and thus better classify the pixel, yielding more accurate segmentation results, e.g., the `Storage tanks' ($2^{nd}$ column) and the `Planes' ($5^{th}$ column). When extended to common natural scenarios, the proposed AgMTR copes equally well, efficiently distinguishing between FG and BG pixels of the query image, as depicted in Fig.\ref{fig:7}. The above advantages sufficiently prove that the proposed AgMTR can efficiently handle the FSS tasks in various scenarios and demonstrate robust generalizability.

\subsection{Ablation Study and Analysis}\label{sec: Ablation Study and Analysis}
A set of ablation studies are executed to discuss and understand our proposed method thoroughly. Unless otherwise noted, the experiments in this section are conducted on iSAID utilizing the DeiT-B/16 backbone under the 1-shot setting.

\begin{table}[t]
\setlength{\abovecaptionskip}{1pt}
\setlength{\belowcaptionskip}{1pt}
\caption{Ablation study of the segmentation performance for the component variants in AgMTR under the 1-shot setting on iSAID. $\Delta$ denotes the performance (mIoU) gain compared to the baseline.} \label{tab:4}
\renewcommand\arraystretch{1.5}
\centering
\resizebox{1.0\linewidth}{!}{
\begin{tabular}{c|ccc|cc|c} 
\toprule
\#  &\renewcommand\arraystretch{1.3} \begin{tabular}[c]{@{}c@{}}Support\\(ALE)\end{tabular} &\renewcommand\arraystretch{1.3} \begin{tabular}[c]{@{}c@{}}Unlabeled\\(AAD)\end{tabular} &\renewcommand\arraystretch{1.3} \begin{tabular}[c]{@{}c@{}}Query\\(SAD)\end{tabular} & mIoU $\uparrow$  & FB-IoU $\uparrow$ & $\Delta$      \\ 
\hline
(a) & \textcolor[rgb]{0.703,0.703,0.703}{\ding{55}}  & \textcolor[rgb]{0.703,0.703,0.703}{\ding{55}}   & \textcolor[rgb]{0.703,0.703,0.703}{\ding{55}}   & 46.82 & 61.12  & -     \\
(b) & \ding{51}  & \textcolor[rgb]{0.703,0.703,0.703}{\ding{55}}   & \textcolor[rgb]{0.703,0.703,0.703}{\ding{55}}   & 48.94 & 62.68  & 2.12  \\
(c) & \ding{51}  & \ding{51}   & \textcolor[rgb]{0.703,0.703,0.703}{\ding{55}}   & 50.01 & \underline{63.04}  & 3.19  \\
(d) & \ding{51}  & \textcolor[rgb]{0.703,0.703,0.703}{\ding{55}}   & \ding{51}   & \underline{50.27} & 62.96  & \underline{3.45}  \\ 
(e) & \ding{51}  & \ding{51}   & \ding{51}   & \textbf{51.58} & \textbf{64.63}  & \textbf{4.76}  \\
\bottomrule
\end{tabular}}
\end{table}

\begin{figure}[t]
\setlength{\abovecaptionskip}{1pt} \setlength{\belowcaptionskip}{1pt}
\centering
\includegraphics[width=1.0\linewidth]{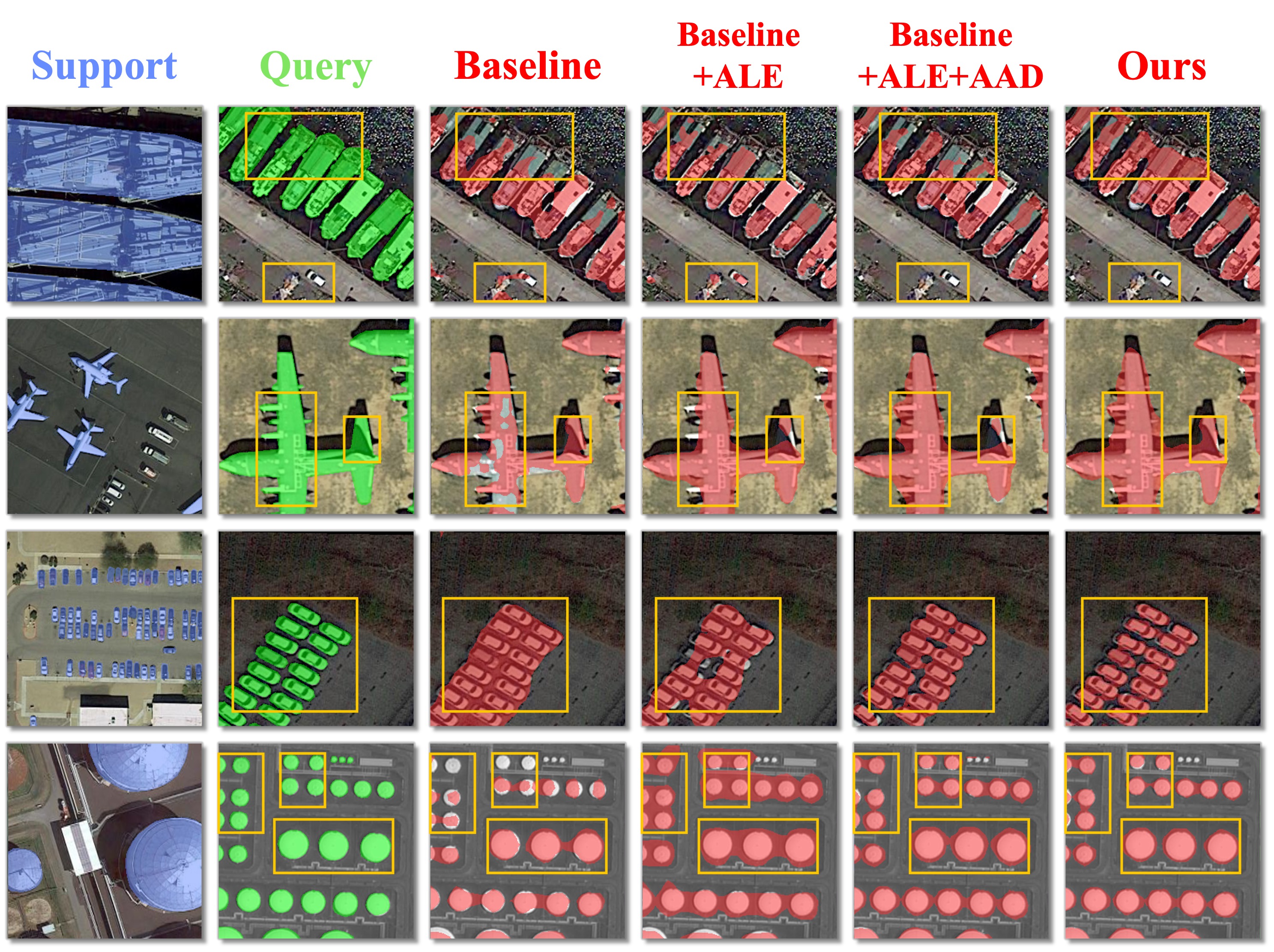}
\caption{\textbf{Qualitative visualization of AgMTR variants and the Baseline.} From left to right, each column represents: Support images (GT), Query images (GT), results with the Baseline, results with the Baseline+ALE, results with the Baseline+ALE+AAD, results with the Baseline+ALE+AAD+SAD (i.e., AgMTR). Several typical areas are marked with yellow boxes.}
\label{fig:8}
\end{figure}

\begin{table}[t]
\setlength{\abovecaptionskip}{1pt}
\setlength{\belowcaptionskip}{1pt}
\caption{Ablation study of the time complexity for the component variants in AgMTR. `GFLOPs' denote the floating point operations of the model and `Inference Speed (FPS)' denotes the number of images the model can infer per second.} \label{tab:4_4}
\renewcommand\arraystretch{1.4}
\centering
\resizebox{1.0\linewidth}{!}{
\begin{threeparttable}
\begin{tabular}{cl|ccc} 
\toprule
\multicolumn{2}{c|}{Method}                  & mIoU          & GFLOPs &\renewcommand\arraystretch{1.1} \begin{tabular}[c]{@{}c@{}}Inference Speed\\(FPS)\end{tabular}  \\ 
\hline
\multicolumn{2}{l|}{IPMT (NeurIPS2021)~\citep{liu2022intermediate}}                    & 48.85          & 123.51                                               & 14.01                                                           \\
\multicolumn{2}{l|}{SCCAN (ICCV2023)~\citep{xu2023self}}                    & 48.31          & 143.37                                               & 16.45                                                           \\
\multicolumn{2}{l|}{BAM (TPAMI2023)~\citep{lang2023base}}                     & 50.43          & 117.03                                               & 30.95                                                           \\ 
\hline
\multirow{6}{*}{\textbf{AgMTR}} & Baseline            & 46.82          & \textbf{47.61}                                       & \textbf{79.71}                                                  \\
& Baseline+ALE         & 48.94          & \underline{48.85}                                        & \underline{71.26}                                                   \\
& Baseline+ALE+AAD     & 50.01          & 111.18                                               & 36.89                                                           \\
& Baseline+ALE+SAD     & 50.27  & 57.59                                                & 41.63                                                           \\
\cline{2-5}
& Baseline+ALE+AAD$^{\dagger}$+SAD & \underline{50.64} & 70.37                                                & 30.30                                                           \\
& Baseline+ALE+AAD+SAD & \textbf{51.58} & 119.93                                                & 28.74                                                           \\

\bottomrule
\end{tabular}
\begin{tablenotes}
\footnotesize
\item[] "ALE" denotes the Agent Learning Encoder, "AAD" denotes the Agent Aggregation Decoder, and "SAD" denotes the Semantic Alignment Decoder. Notably, "AAD" employs 5 unlabeled images by default, while "AAD$^\dagger$" employs 1 unlabeled image.
\end{tablenotes}
\end{threeparttable}}
\end{table}

\begin{figure}[t]
\setlength{\abovecaptionskip}{1pt} \setlength{\belowcaptionskip}{1pt}
\centering
\includegraphics[width=1.0\linewidth]{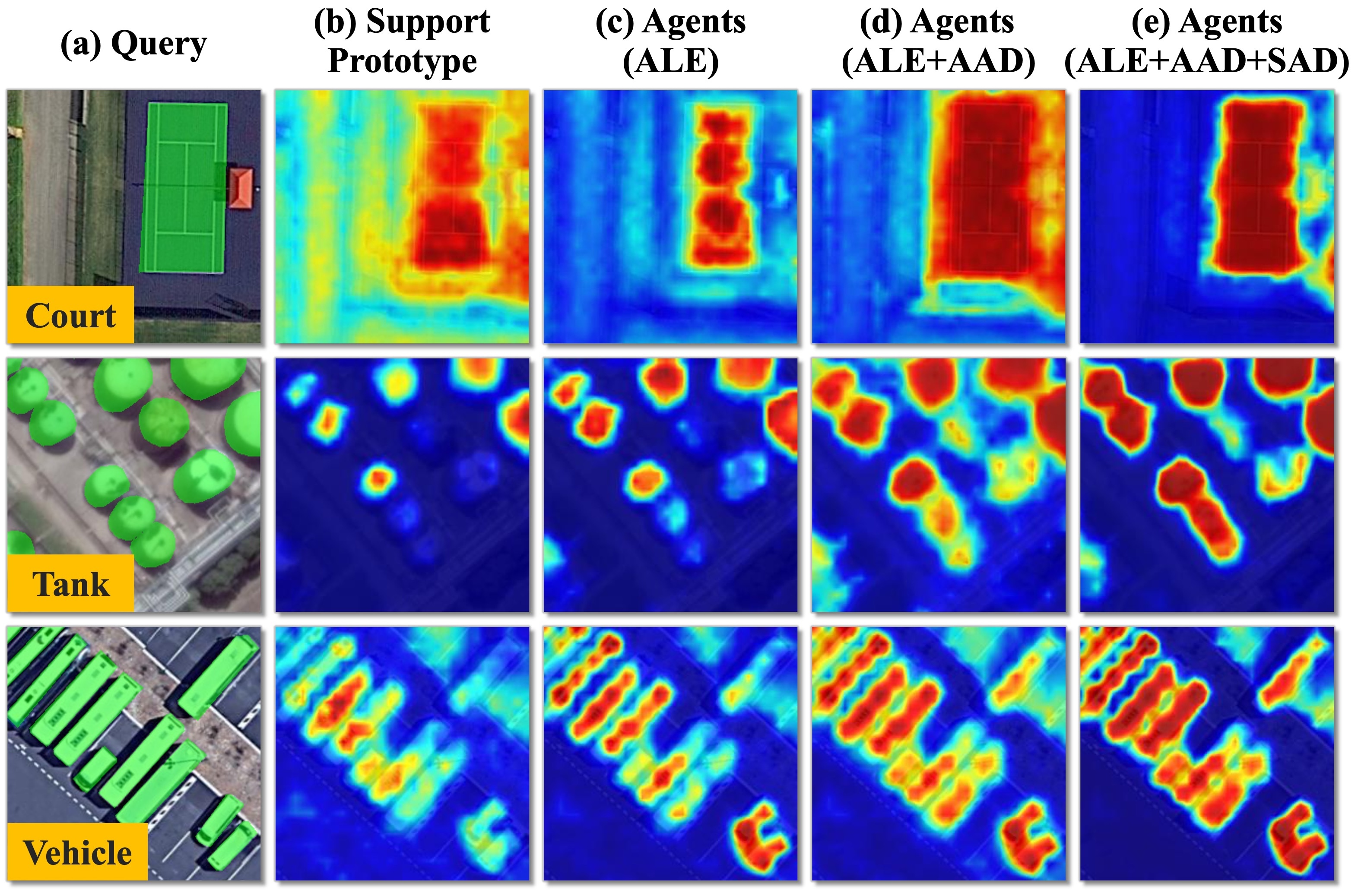}
\caption{\textbf{Qualitative visualization of activation maps of component variants in AgMTR on the query image.} As can be noticed, each component promotes the agent better to activate the specific objects in the query image.}
\label{fig:11_1}
\end{figure}

\begin{table}[t]
\setlength{\abovecaptionskip}{1pt}
\setlength{\belowcaptionskip}{1pt}
\caption{Ablation study of the intra-class variance for the component variants in AgMTR by measuring the cosine distance $D_{aq}$ between the agents and the query prototype on iSAID dataset, where the distance $D_{sq}$ between the support prototype and the query prototype is employed as a comparison.} \label{tab:5}
\renewcommand\arraystretch{1.8}
\centering
\resizebox{1.0\linewidth}{!}{
\begin{tabular}{c|ccc|ccc|c} 
\toprule
\multirow{2}{*}{\#}   & \multicolumn{3}{c|}{Agent Semantic Source} & \multicolumn{4}{c}{Cosine distance $\uparrow$}      \\ 
\cline{2-8}
&\renewcommand\arraystretch{1.3} \begin{tabular}[c]{@{}c@{}}Support\\(ALE)\end{tabular} &\renewcommand\arraystretch{1.3} \begin{tabular}[c]{@{}c@{}}Unlabeled\\(AAD)\end{tabular} &\renewcommand\arraystretch{1.3} \begin{tabular}[c]{@{}c@{}}Query\\(SAD)\end{tabular} & Fold-0 & Fold-1 & Fold-2 & Mean   \\ 
\hline
$D_{sq}$                  &\textcolor[rgb]{0.703,0.703,0.703}{\ding{55}}         &\textcolor[rgb]{0.703,0.703,0.703}{\ding{55}}           &\textcolor[rgb]{0.703,0.703,0.703}{\ding{55}}                & 0.5097 & 0.5357 & 0.7076 & 0.5843  \\ 
\hline
\multirow{3}{*}{$D_{aq}$} & \ding{51}        &\textcolor[rgb]{0.703,0.703,0.703}{\ding{55}}           &\textcolor[rgb]{0.703,0.703,0.703}{\ding{55}}                & 0.6778 & 0.6437 & 0.7310 & 0.6842  \\
&\ding{51}         &\ding{51}           &\textcolor[rgb]{0.703,0.703,0.703}{\ding{55}}                & 0.7076 & 0.6498 & 0.7614 & 0.7063  \\
&\ding{51}         &\ding{51}           &\ding{51}                & \textbf{0.7678} & \textbf{0.7047} & \textbf{0.7669} & \textbf{0.7465}  \\
\bottomrule
\end{tabular}}
\end{table}
\subsubsection{Component Analysis}\label{sec:Component Analysis}

\textbf{Component Analysis in Segmentation Performance:}
Table \ref{tab:4} discusses the performance impact of the component variants in AgMTR. In general, we could observe the following phenomena: (i) With the integration of Agent Learning Encoder (ALE), Agent Aggregation Decoder (AAD), and Semantic Alignment Decoder (SAD) three components, AgMTR achieves a 4.76\% mIoU improvement compared to the baseline, strongly validating their effectiveness. (ii) The proposed ALE contributes the most to the final model. This is possible because mining class-specific semantics from the support images is the most essential and fundamental step. Only when sufficiently valid semantics are mined can agents stably mine interested semantics from unlabeled and query images without introducing irrelevant noise. This also validates the effectiveness of the proposed agent paradigm to some extent. (iii) in case (c) and case (d), the introduction of AAD and SAD yields 1.07\% and 1.33\% gains, respectively, which quantitatively validates that the aggregation of class-specific semantics from unlabeled and query images for agents can facilitate breaking through the limited support set and better guide query segmentation. (IV) Besides, we present the segmentation results for different variants as shown in Fig.\ref{fig:8}. By introducing components step by step, the segmentation results are further improved, qualitatively verifying the effectiveness and necessity of the proposed components.

\noindent \textbf{Component Analysis in Complexity:}
To investigate the time complexity of the different components in AgMTR, we conduct an ablation study on the iSAID dataset, with the results depicted in Table \ref{tab:4_4}. The following phenomena can be derived:
(i) Among the three components, AAD has the largest computational complexity. This is because AAD aims to mine class-specific information from additional unlabeled images (5 images are utilized by default), which equates to processing 7 (1 query, 1 support, and 5 unlabeled) images simultaneously, incurring additional computational costs.
(ii) Despite the obvious computational cost of AAD, AgMTR achieves the best mIoU performance with the second-lowest GFLOPs compared to others, far better than SCCAN, i.e., 51.58\% vs 48.31\% mIoU and 119.93 vs 143.37 GFLOPs.
(iii) AgMTR achieves a balance between accuracy and complexity when AAD utilizes less unlabeled images. We attempt to mine class-specific semantics with only one unlabeled image, at which point the model achieves 50.64\% mIoU at 70.37 GFLOPs, balancing segmentation accuracy and computational complexity.

\noindent \textbf{Component Analysis in Intra-class Variance:}
To objectively evaluate the effectiveness of the proposed three components, we employ cosine distance to measure the distance $D_{aq}$ between the query prototype and the agents in every episode, where the distance between the query prototype and the support prototype $D_{sq}$ serves as a comparison. The average distance of the classes in each fold and the average of all classes are listed in Table \ref{tab:5}. Notably, multiple agents are averaged into one for computation with the query prototype. Obviously, after gradually aggregating the semantics from the support images, the unlabeled images, and the query image by driving the agents through ALE, AAD, and SAD, $D_{aq}$ gradually increases, and is larger than $D_{sq}$ in all folds. This phenomenon reveals that with the proposed three components, the proposed agents work significantly to reduce the distance with the query image and minimize the negative impact of large intra-class variations in remote sensing, providing better guidance for query image segmentation. We also visualize the activation maps of the agent derived from different components, as illustrated in Fig.\ref{fig:11_1}. Compared to the support prototype, agents derived from different components can focus more on specific objects in the query image, enabling more precise activations. The above results fully validate the effectiveness of the three proposed components from the perspective of intra-class variance.

\subsubsection{Ablation study of the Agent Learning Encoder}\label{sec:Ablation study of the Agent Learning Encoder (ALE)}
As mentioned in Sec.\ref{sec:Agent Learning Encoder}, Agent Learning Encoder (ALE) aims to mine class-specific semantics in the support images to derive local-aware agents for performing agent-level semantic correlation. The gain of 2.12\% mIoU in Table \ref{tab:4} has quantitatively verified that the local agents generated by utilizing ALE are successful in effectively distinguishing between foreground and background pixels, generating more complete objects and more accurate boundaries (see Fig.\ref{fig:8}). This section discusses specific details in the ALE, including the number of agents generated and the learning scheme.

\begin{figure}[t]
\setlength{\abovecaptionskip}{1pt}
\setlength{\belowcaptionskip}{1pt}
\centering
\includegraphics[width=1.0\linewidth]{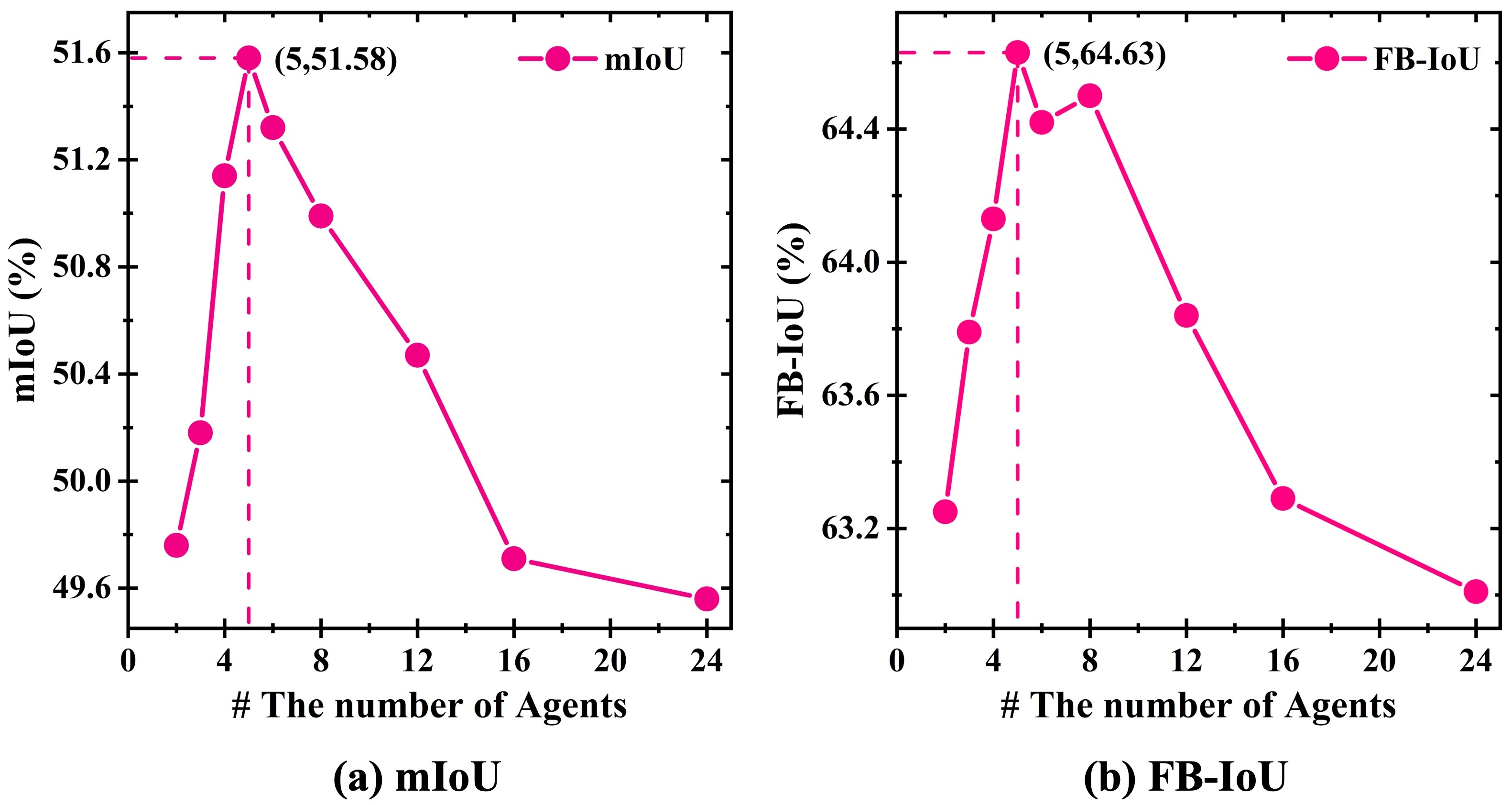}
\caption{\textbf{Ablation study of the number of agents (i.e., $N_a$)} on segmentation performance. The best segmentation accuracy is obtained with the number of agents set to 5, i.e., $N_a=5$.}
\label{fig:11}
\end{figure}

Firstly, we investigate the impact of the number of agents (i.e., $N_a$) on segmentation performance, as depicted in Fig.\ref{fig:11}. It can be found that as the number of agents (i.e., $N_a$) increases from 1 to 5, the segmentation accuracy will gradually increase, especially with 51.58\% mIoU achieved for $N_{a}=5$. It is reasonable because the support mask will be divided into more delicate local regions, thus driving the agent to aggregate finer-grained local semantics. However, when continuing to increase $N_a$, the segmentation accuracy begins to decrease. At this point, the support mask becomes forced to be divided into more unreasonable local masks, such as dividing the left and right `Wings' of the `Plane', which leads to different agents aggregating similar semantics, resulting in redundancy and even interference. 

\begin{table}[t]
\setlength{\abovecaptionskip}{1pt}
\setlength{\belowcaptionskip}{1pt}
\caption{Ablation study of different initialization and learning schemes of agents in ALE.} \label{tab:7}
\renewcommand\arraystretch{1.6}
\centering
\resizebox{1.0\linewidth}{!}{
\begin{tabular}{c|c|c|ccc|cc} 
\toprule
\#&Initialization & OT & Fold-0 & Fold-1 & Fold-2 & Mean  & FB-IoU  \\ 
\hline
(a) &Random                 & \ding{51}  & 52.50   & 41.87  & 53.00     & 49.12 & 63.44   \\
(b) &Random+Sup             & \textcolor[rgb]{0.703,0.703,0.703}{\ding{55}}  & 55.64  & 42.36  & 54.22  & 50.74 & 63.82   \\
(c) &Random+Sup             & \ding{51}  & 57.51  & 43.54  & 53.69  & \textbf{51.58} & \textbf{64.63}   \\
\bottomrule
\end{tabular}}
\end{table}

\begin{figure}[t]
\setlength{\abovecaptionskip}{1pt}
\setlength{\belowcaptionskip}{1pt}
\centering
\includegraphics[width=1.0\linewidth]{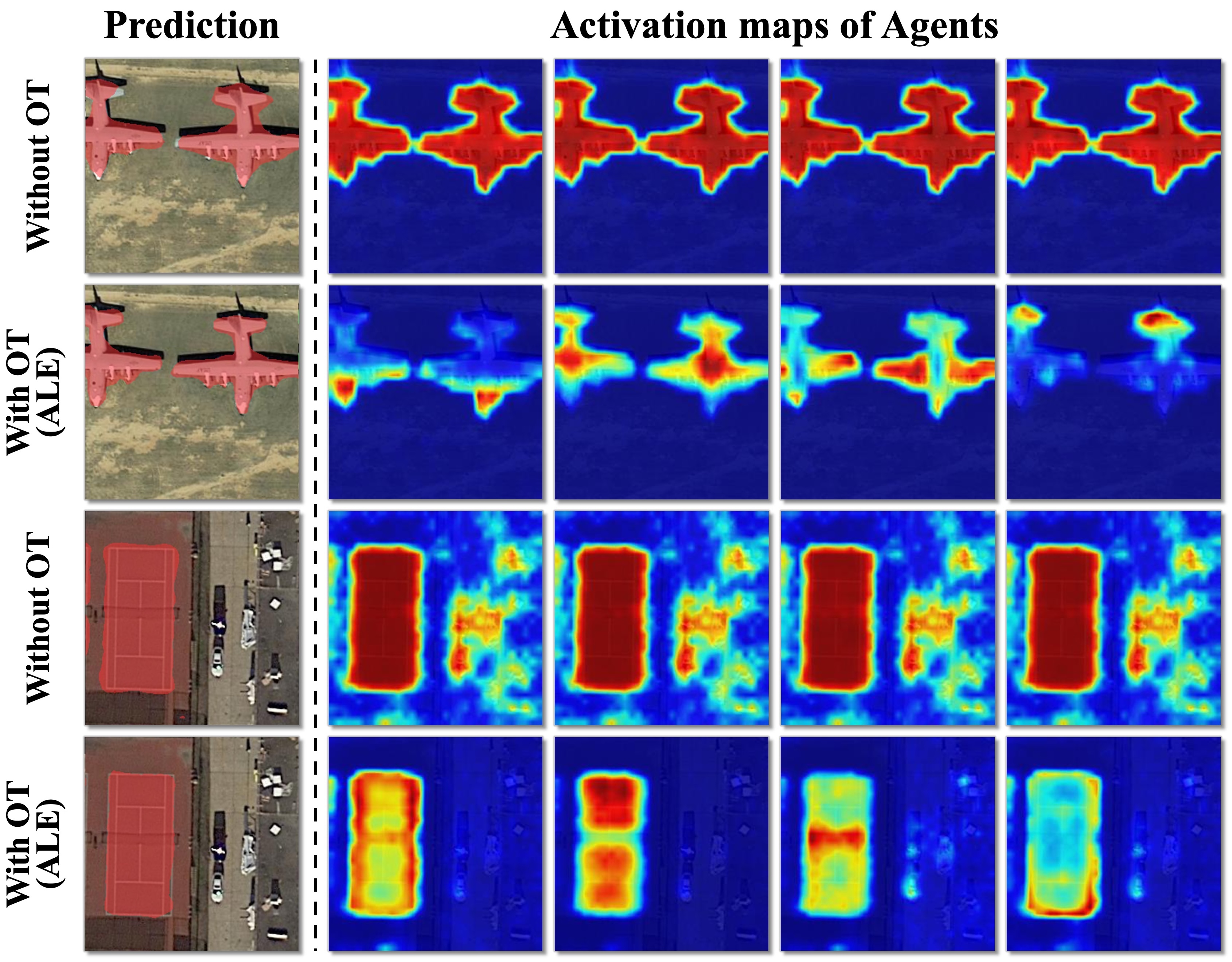}
\caption{\textbf{Qualitative visualization of activation maps of different agents on the query image}, where `with/without OT' denotes whether ALE employs OT to generate multiple local masks.}
\label{fig:11_4}
\end{figure}

We then explore the performance impact of different initialization and learning schemes of agents in ALE, as illustrated in Table \ref{tab:7}. In these schemes, `Random' and `Random+Sup' denote randomly initialized tokens, and the addition of the support prototype with randomly initialized tokens as initial agents, respectively. `OT' denotes dividing the support mask into local masks with the optimal transport algorithm. From the table, we can observe the following phenomena: (i) Utilizing only randomly initialized tokens as the initial agents is suboptimal (case (a)) since the agents lack class-specific semantics to guide the capture of similar object semantics. Instead, dependent on class-specific semantics in the support prototype (case (c)), the agents can continue to assimilate similar semantics in subsequent processes, thus further enhancing class-awareness. (ii) After introducing the optimal transport (OT) to adaptively generate local masks, ALE can drive different agents to aggregate the interested semantics under different local masks. The comparison between case (b) and case (c) proves the effectiveness of the above scheme. Fig.\ref{fig:11_4} also visualizes the activation maps of different agents on the query image after generating local masks employing optimal transport (OT). It can be found that equipped with OT, ALE can drive different agents to focus on different local regions of the specific object. When OT is not employed, not only the generated agents do not have local-awareness and produce coarser activations, but also different agents will aggregate similar semantics and produce redundancy. The above phenomenon effectively proves the effectiveness and necessity of employing OT to explore local semantics in ALE.

\begin{figure}[t]
\setlength{\abovecaptionskip}{1pt} 
\setlength{\belowcaptionskip}{1pt}
\centering
\includegraphics[width=1.0\linewidth]{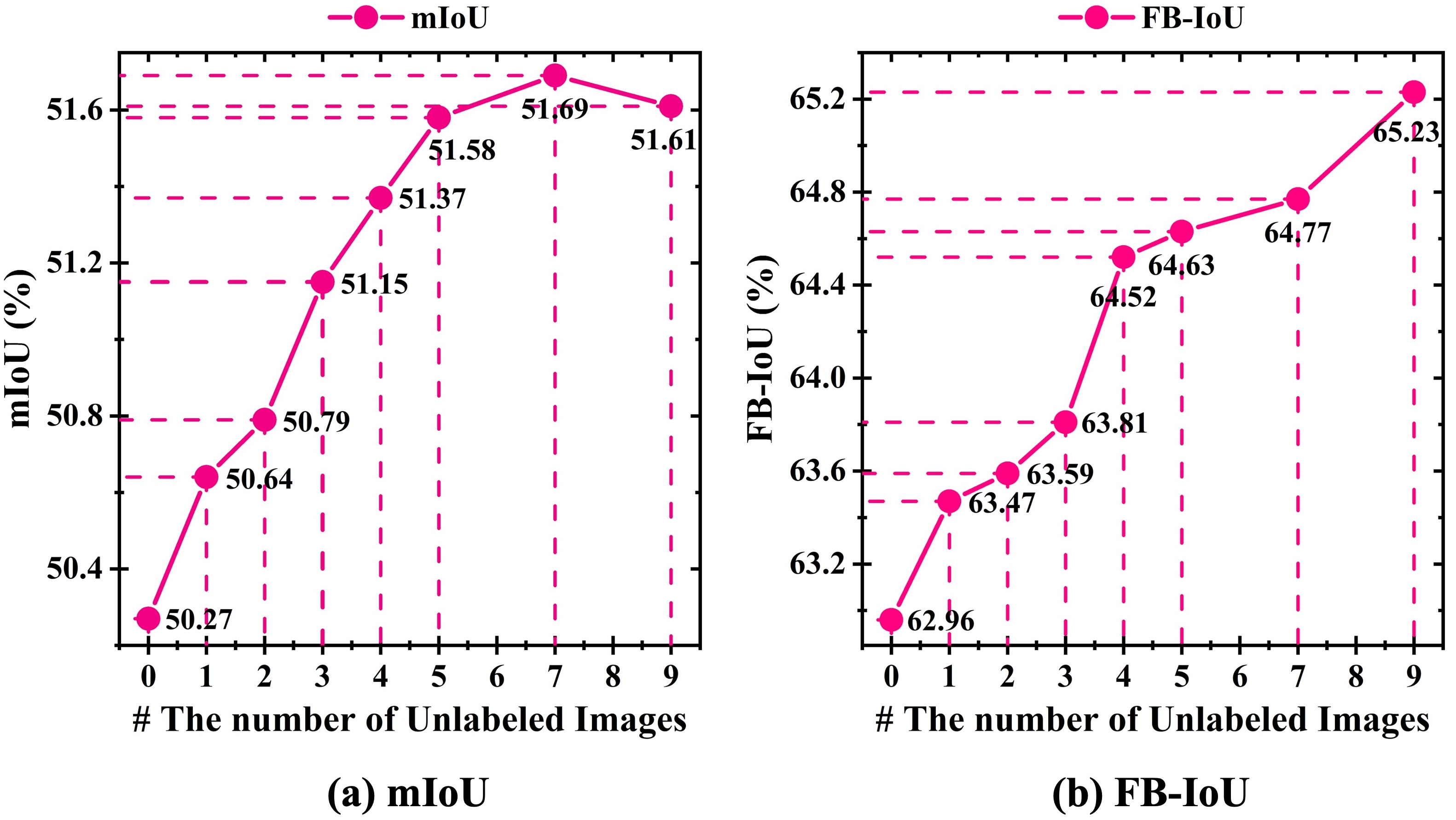}
\caption{\textbf{Ablation study of the number of unlabeled images (i.e., $N_u$)} on segmentation performance in AAD. As the number increases, the segmentation accuracy gradually improves, especially for FB-IoU.}
\label{fig:12}
\end{figure}

\begin{figure}[t]
\setlength{\abovecaptionskip}{1pt} \setlength{\belowcaptionskip}{1pt}
\centering
\includegraphics[width=1.0\linewidth]{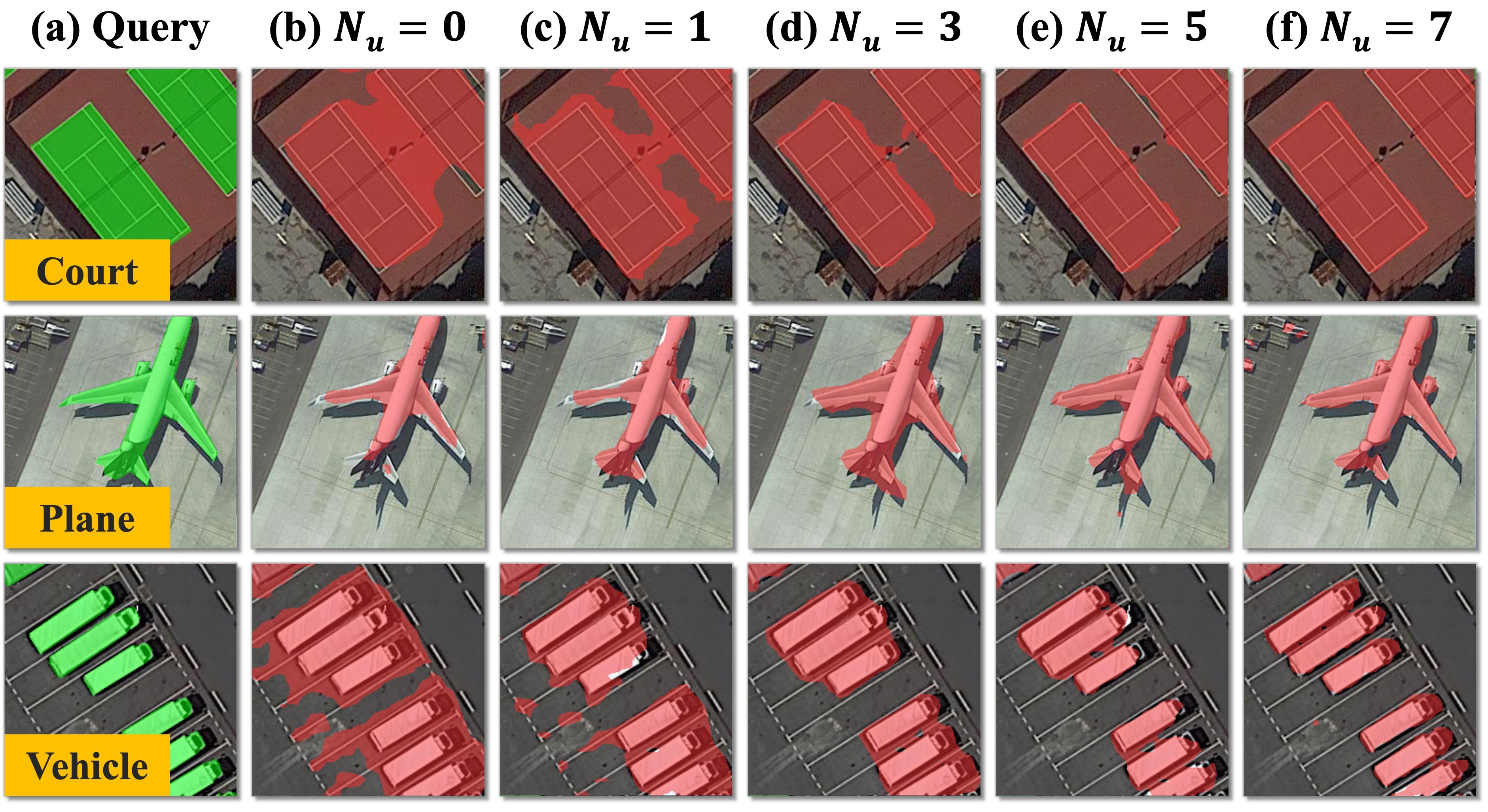}
\caption{\textbf{Qualitative visualization with different numbers of unlabeled images in AAD.}}
\label{fig:13_1}
\end{figure}

\subsubsection{Ablation study of the Agent Aggregation Decoder}\label{sec:Ablation study of the Agent Aggregation Decoder (AAD)}

As mentioned in Sec.\ref{sec: Agent Aggregation Decoder}, Agent Aggregation Decoder (AAD) aims to mine semantic information in unlabeled images that are beneficial to the current task, for breaking the limitations of the support set and enabling stronger generalization. The quantitative results in Table \ref{tab:5} demonstrate that AAD can effectively mine class-specific semantics in unlabeled data sources, flexibly dealing with intra-class variance. A natural discussion is about how many unlabeled images (i.e., $N_u$) are the most appropriate. Fig.\ref{fig:12} explores the impact of the number of unlabeled images (i.e., $N_u$) on segmentation performance. As the number of unlabeled images increases, AAD can mine richer class-specific semantics from the larger number of unlabeled images, thus further optimizing the agents to break through the limited support set and better guide query segmentation.The above observation is also qualitatively demonstrated by the segmentation results with different numbers of unlabeled images in Fig.\ref{fig:13_1}. Considering the efficiency of the model (see Table \ref{tab:8_1}), $N_u$ is set to 5 in the experiments. We then investigate the impact of Graph Attention Network (GAT) on performance in AAD, as illustrated in Table \ref{tab:8}. It can be found that 0.69\% mIoU and 0.87\% FB-IoU improvement is realized after introducing GAT, which is a good demonstration that GAT can enhance the contextual semantics of unlabeled prototypes and thus be efficiently captured by local agents.

\begin{table}[t]
\setlength{\abovecaptionskip}{1pt}
\setlength{\belowcaptionskip}{1pt}
\caption{Ablation study of the number of unlabeled images (i.e., $N_u$) on computational complexity measured in GFLOPs.} \label{tab:8_1}
\renewcommand\arraystretch{1.4}
\centering
\resizebox{1.0\linewidth}{!}{
\begin{tabular}{c|cccccccc} 
\toprule
$N_{u}$ & 0     & 1     & 2     & 3     & 4     & 5      & 7      & 9       \\ 
\hline
mIoU    & 50.27 & 50.64 & 50.79 & 51.15 & 51.37 & 51.58  & 51.69  & 51.61   \\
GFLOPs  & 57.59 & 70.37 & 82.92 & 95.46 & 107.57 & 119.93 & 143.77 & 166.83  \\
\bottomrule
\end{tabular}}
\end{table}

\begin{table}[t]
\setlength{\abovecaptionskip}{1pt}
\setlength{\belowcaptionskip}{1pt}
\caption{Ablation study of the Graph Attention Network (GAT) in AAD.} \label{tab:8}
\renewcommand\arraystretch{1.4}
\centering
\resizebox{0.9\linewidth}{!}{
\begin{tabular}{c|c|ccc|cc} 
\toprule
\#&GAT & Fold-0 & Fold-1 & Fold-2 & Mean  & FB-IoU  \\ 
\hline
(a)&\textcolor[rgb]{0.703,0.703,0.703}{\ding{55}}   & 56.22  & 41.61  & 54.83  & 50.89 & 63.76   \\
(b)&\ding{51}   & 57.51  & 43.54  & 53.69  & \textbf{51.58} & \textbf{64.63}   \\
\bottomrule
\end{tabular}}
\end{table}

\begin{figure}[t]
\setlength{\abovecaptionskip}{1pt} \setlength{\belowcaptionskip}{1pt}
\centering
\includegraphics[width=1.0\linewidth]{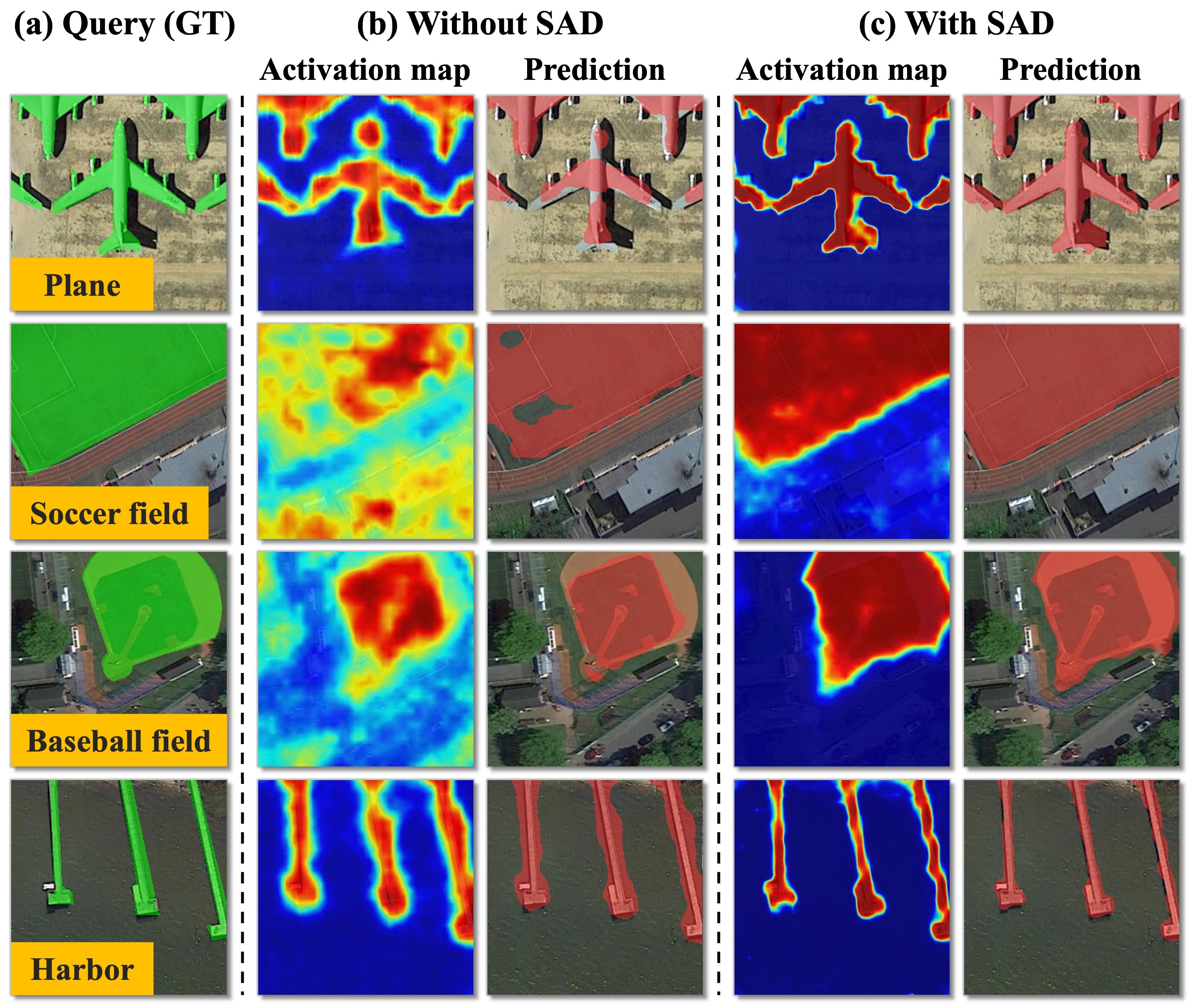}
\caption{\textbf{Qualitative visualization of Semantic Alignment Decoder (SAD).} (a) Query image with GT. (b) Activation map and prediction result without SAD. (c) Activation map and prediction result with SAD. }
\label{fig:13_2}
\end{figure}

\begin{figure}[t]
\setlength{\abovecaptionskip}{1pt} \setlength{\belowcaptionskip}{1pt}
\centering
\includegraphics[width=1.0\linewidth]{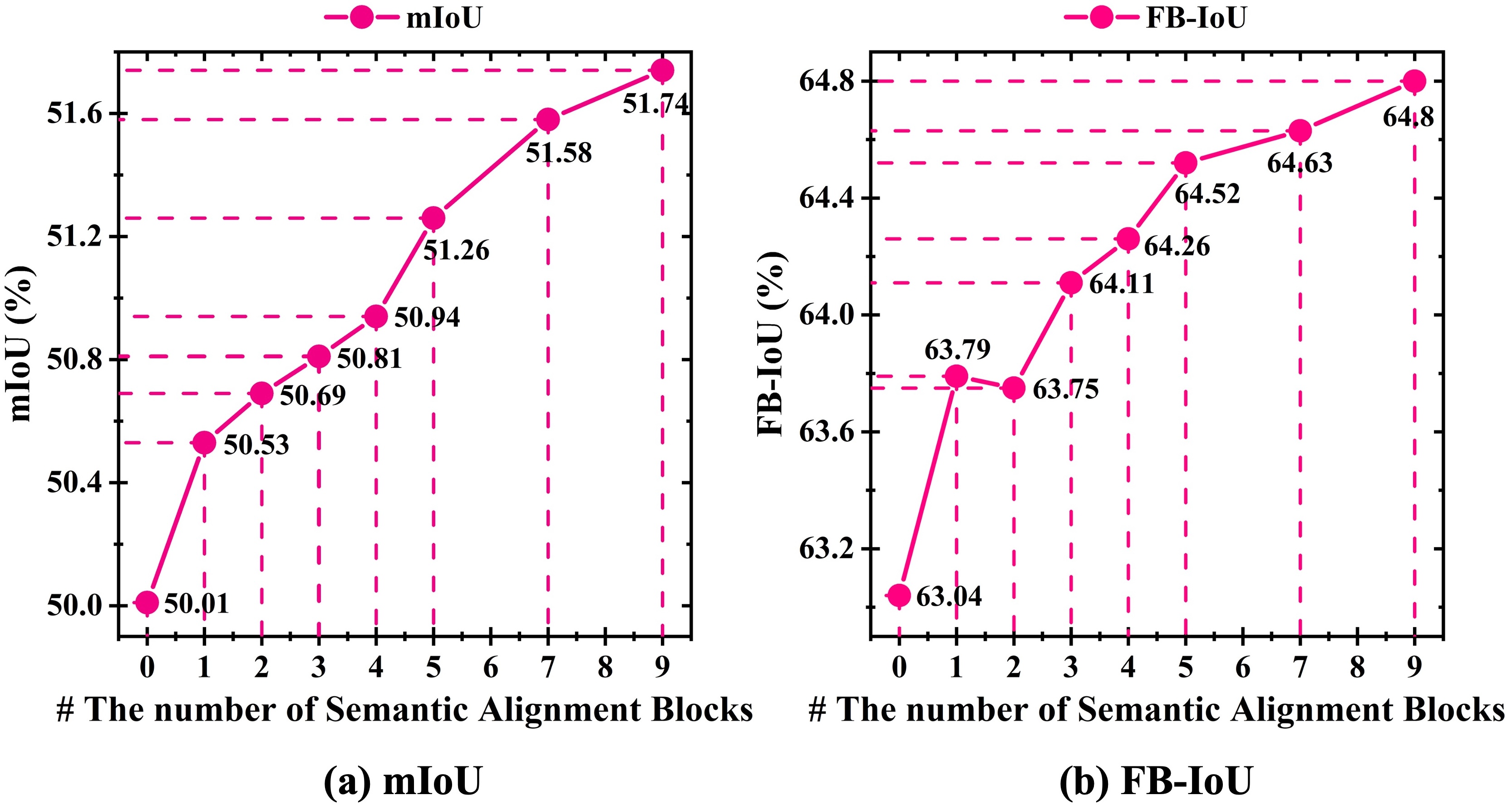}
\caption{\textbf{Ablation study of the number of Semantic Alignment Blocks (i.e., $N$)} on segmentation performance in SAD.}
\label{fig:13}
\end{figure}

\begin{table}[t]
\setlength{\abovecaptionskip}{1pt}
\setlength{\belowcaptionskip}{1pt}
\caption{Ablation study of the contribution $\gamma$ of Agent Segmentation Loss $\mathcal{L}_{ASL}$ to total loss in Eq.(\ref{equation:15}).} \label{tab:9}
\renewcommand\arraystretch{1.4}
\centering
\resizebox{0.9\linewidth}{!}{
\begin{tabular}{c|cccccc} 
\toprule
$\mathcal{L}_{ASL}$ & 0.0     & 0.2   & 0.4   & 0.6   & 0.8   & 1.0      \\ 
\hline
mIoU   & 51.30  & 51.19 & 51.27 & 51.45 & \textbf{51.58} & 51.51  \\
FB-IoU & 64.46 & 64.45 & 64.07 & 64.26 & \textbf{64.63} & 64.39  \\
\bottomrule
\end{tabular}}
\end{table}

\subsubsection{Ablation study of the Semantic Alignment Decoder}\label{sec:Ablation study of the Semantic Alignment Decoder (SAD)}

As mentioned in Sec.\ref{sec:Semantic Alignment Decoder}, Semantic Alignment Decoder (SAD) aims to mine the class-specific semantics of the query image itself to promote semantic alignment with the query object, and this paradigm of mining one's own semantics to guide one's own segmentation can cope well with large intra-class variations. We explore the distance with the query object in feature space, with the results shown in Table \ref{tab:5}, which quantitatively demonstrates the effectiveness of SAD in bringing the distance with the query object closer. The visualization results in Fig.\ref{fig:13_2} also qualitatively validate that the proposed SAD can further mine the query image class-specific semantics to achieve more precise boundaries and fewer false activations.

Further, we explore the impact of the number of Semantic Alignment Blocks (i.e., $N$) on segmentation performance, as depicted in Fig.\ref{fig:13}. As the number $N$ increases, SAD can further improve the quality of the pseudo-local masks of the query image, which enables the agents to better aggregate the local semantics from the query image, achieving better segmentation, especially receiving 51.74\% mIoU and 64.80\% FB-IoU when $N=9$. Considering the efficiency of the model, $N=7$ is a proper choice. In addition, Table \ref{tab:9} gives the impact of the contribution $\gamma$ of the Agent Segmentation Loss $\mathcal{L}_{ASL}$ in SAD. It can be observed that the segmentation accuracy shows an increasing and then decreasing trend, and reaches the best performance when $\gamma=0.8$.

\begin{figure}[t]
\setlength{\abovecaptionskip}{1pt} \setlength{\belowcaptionskip}{1pt}
\centering
\includegraphics[width=1.0\linewidth]{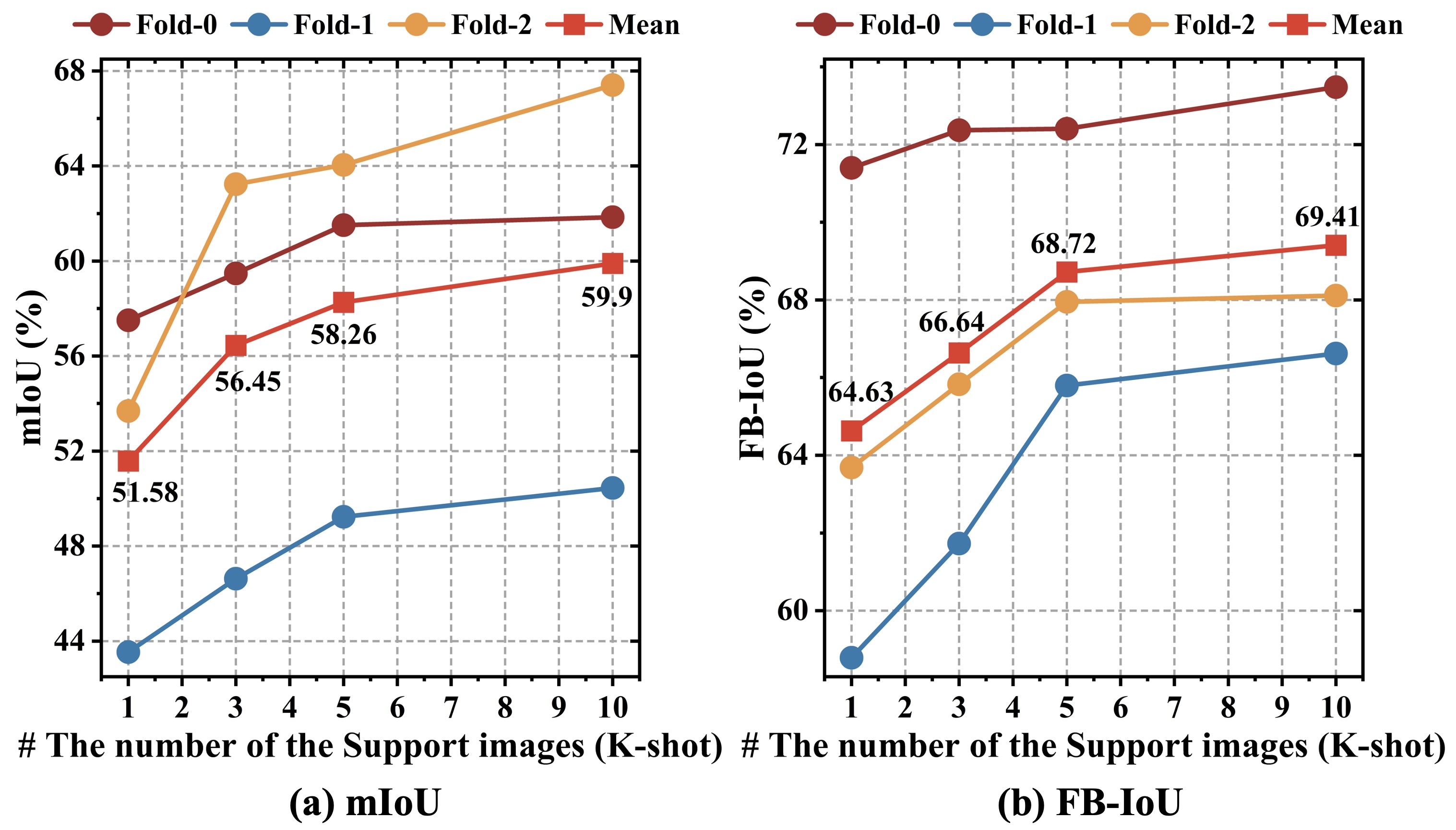}
\caption{\textbf{Ablation study of the number of support images (i.e., K-shot)} on segmentation performance.}
\label{fig:kshot}
\end{figure}

\begin{figure}[t]
\setlength{\abovecaptionskip}{1pt} \setlength{\belowcaptionskip}{1pt}
\centering
\includegraphics[width=1.0\linewidth]{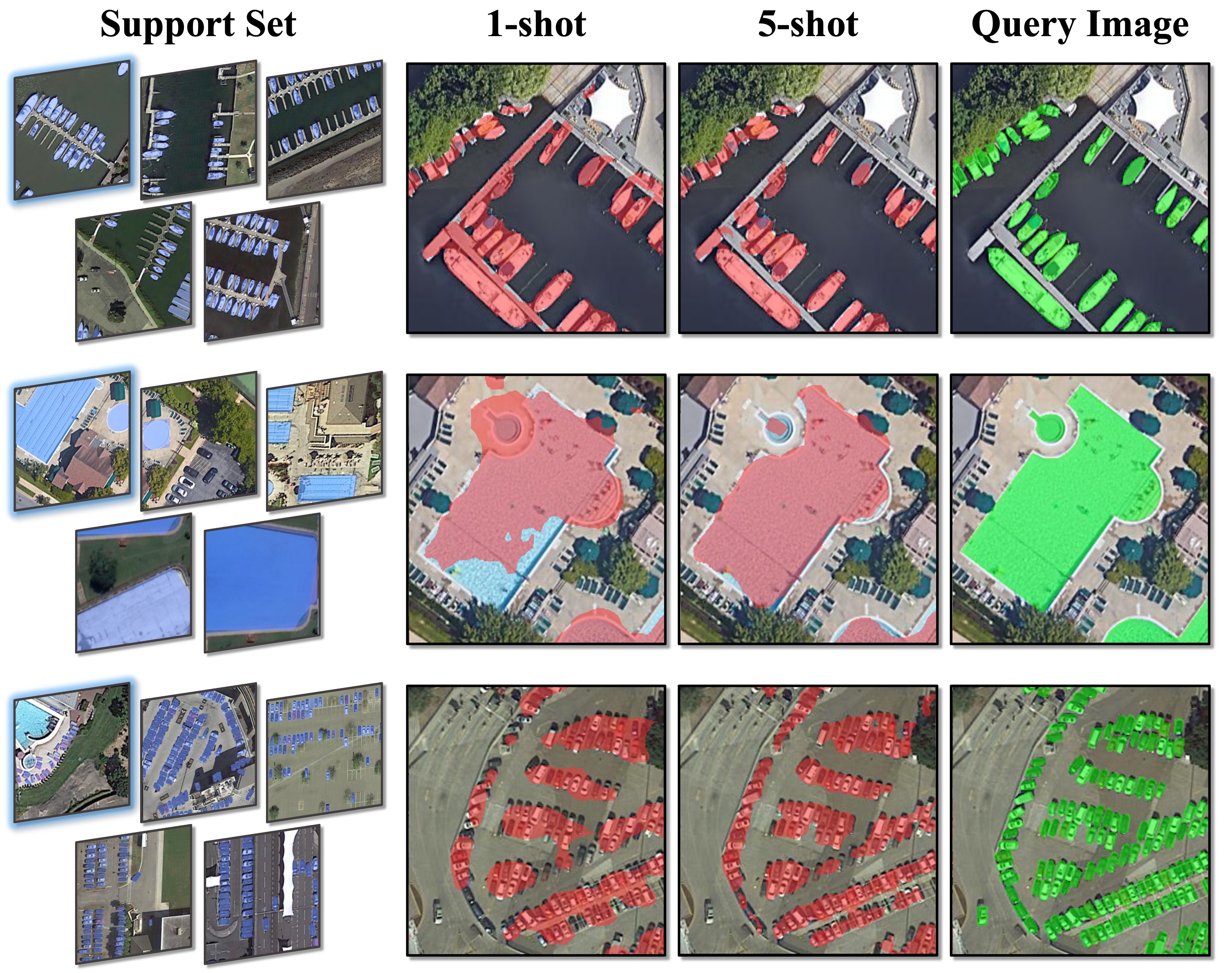}
\caption{\textbf{Comparison results of our method under the 1- and 5-shot settings on iSAID dataset.} The first labeled image (blue border) in the Support Set is sampled for 1-shot segmentation.}
\label{fig:vis_kshot}
\end{figure}

\subsubsection{Ablation study of K-shot}\label{sec:Ablation study of K-shot}
When expanded to the K-shot (K$>$1) setting, the support set can provide more labeled samples for guiding segmentation. Fig.\ref{fig:kshot} demonstrates the impact on performance in iSAID with different numbers of support images provided. It can be noticed that the mIoU of each Fold achieves a large improvement as $K$ keeps increasing. When continuing to increase to 10 images, the boost slows down, which is reasonable because of the decreasing marginal effect of the labeled samples. Fig.\ref{fig:vis_kshot} also compares the segmentation results of the proposed AgMTR under the 1- and 5-shot settings. The 5-shot results are significantly superior to the 1-shot results with more complete objects and fewer false activations. The insight behind this phenomenon is that AgMTR can mine more class-representative and stable agents from more support images, which can further facilitate agents to capture similar semantics from unlabeled and query images for more precise agent-level semantic correlations.

\begin{figure}[t]
\setlength{\abovecaptionskip}{1pt} \setlength{\belowcaptionskip}{1pt}
\centering
\includegraphics[width=1.0\linewidth]{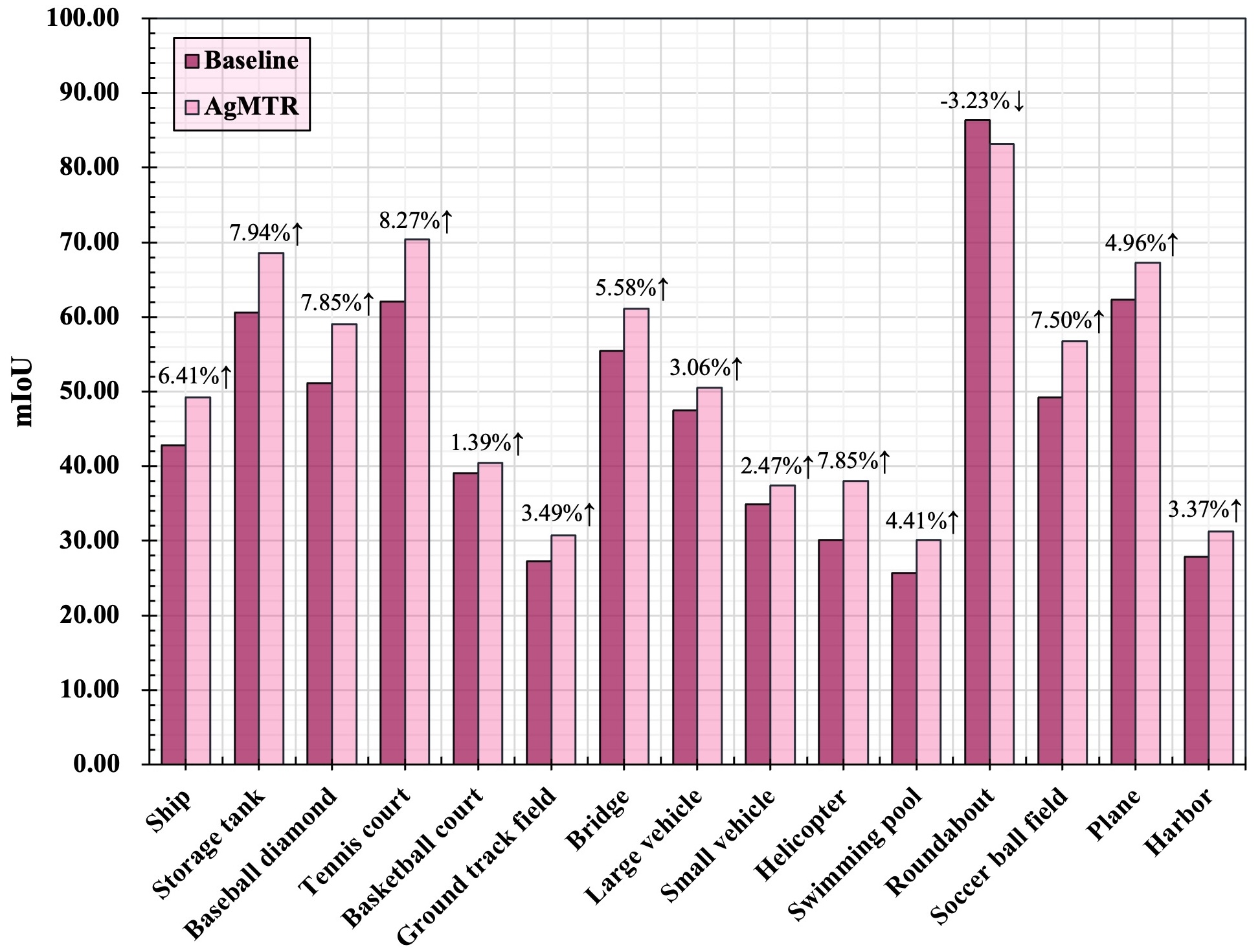}
\caption{\textbf{Performance comparison (mIoU) of our AgMTR and the Baseline} in different classes on iSAID under the 1-shot setting.}
\label{fig:9}
\end{figure}

\begin{figure*}[t]
\setlength{\abovecaptionskip}{1pt} \setlength{\belowcaptionskip}{1pt}
\centering
\includegraphics[width=1.0\linewidth]{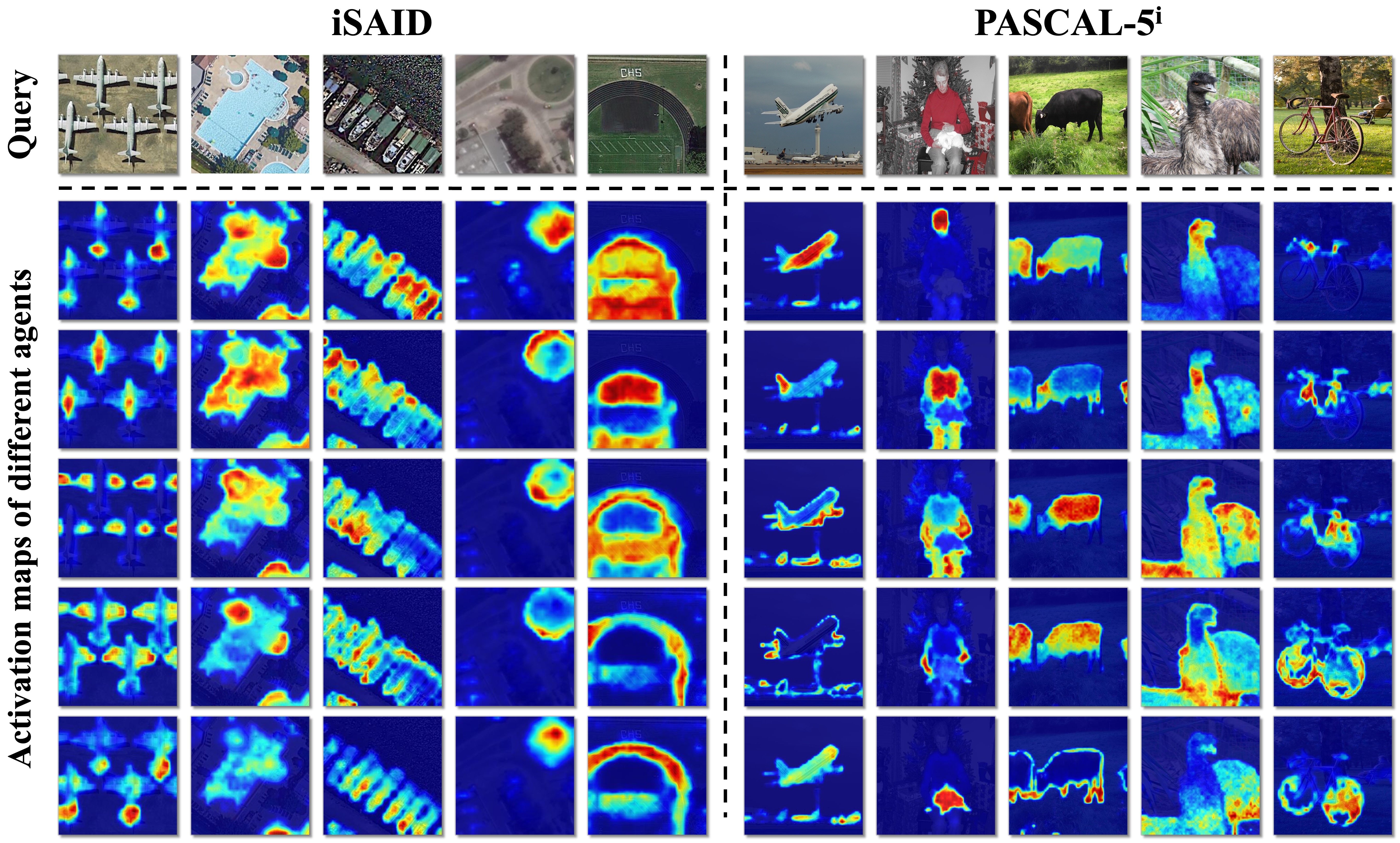}
\caption{\textbf{Qualitative visualization of activation maps of the proposed agents on the query image in iSAID and PASCAL-5$^i$.} Obviously, different agents reasonably focus on different local regions. This is because the proposed three components could explore valuable local semantics from the support, unlabeled, and query images to empower the agents with local-awareness.}
\label{fig:10}
\end{figure*}

\subsubsection{Performance Analysis for Each Class}\label{sec:Performance Analysis for Each Class}
Fig.\ref{fig:9} compares the segmentation performance of our AgMTR with the baseline in different classes on iSAID. Our method achieves significant advantages in 14 out of 15 classes, especially for the `Storage tank', `Baseball field', `Tennis court', `Helicopter', and `Soccer field' classes achieving more than 7\% gains. We observe that these classes tend to have large intra-class differences, requiring the FSS model to be more class-aware. However, our method performs worse than the baseline for the `Roundabout' class, i.e., 83.14\% $vs$ 85.27\%. It may be due to the fact that the appearance of this class tends to be regular and no other classes are interfering in the image, the baseline method can easily tackle it as well.

\begin{table}[t]
\setlength{\abovecaptionskip}{1pt}
\setlength{\belowcaptionskip}{1pt}
\caption{Ablation study of the Backbone in the proposed AgMTR, where the inference speed is measured under the 5-shot setting.} \label{tab:10}
\renewcommand\arraystretch{1.7}
\centering
\resizebox{1.0\linewidth}{!}{
\begin{tabular}{c|c|cc|cc|c} 
\toprule
\multirow{2}{*}{\#} & \multirow{2}{*}{Backbone} & \multicolumn{2}{c|}{1-shot}     & \multicolumn{2}{c|}{5-shot}     &\renewcommand\arraystretch{1.3} \multirow{2}{*}{\begin{tabular}[c]{@{}c@{}}Inference Speed \\(FPS)\end{tabular}}  \\ 
\cline{3-6}
&                           & mIoU           & FB-IoU         & mIoU           & FB-IoU         &                                                                                   \\ 
\hline
(a)                 & ViT-S/16                  & 48.70          & 61.79          & 55.45          & 66.57          & \textbf{25.02}                                                                    \\
(b)                 & ViT-B/16                  & \underline{49.98}  & \underline{63.20}  & \underline{57.00}  & \underline{67.62}  & {18.45}                                                                     \\ 
\hline
(c)                 & DeiT-T/16                 & 46.84          & 61.28          & 54.70          & 65.99          & \underline{23.91}                                                                             \\
(d)                 & DeiT-S/16                 & 49.11          & 62.52          & 55.33          & 66.31          & 22.49                                                                             \\
(e)                 & DeiT-B/16                 & \textbf{51.58} & \textbf{64.63} & \textbf{58.26} & \textbf{68.72} & 19.68                                                                             \\
\bottomrule
\end{tabular}}
\end{table}

\subsubsection{Ablation study of the Backbone}\label{sec:Ablation study of the backbone}
Table \ref{tab:10} gives the impact of different backbones on segmentation performance, where the inference speed is measured under the 5-shot setting. We can observe the following phenomena: (i) Employing the backbone of DeiT-B/16 realizes the optimal accuracy in all settings, followed by the backbone of ViT-B/16. (ii) Employing the backbone of DeiT-S/16 (89MB) yields a segmentation accuracy on par with the ResNet-101 (179MB) based method DMNet, i.e., 49.11\% $vs$ 49.21\%, which clearly demonstrates the superiority of the proposed method. (iii) ViT-S/16 and DeiT-T/16 have slightly worse segmentation performance, but they have faster inference speeds than others.

\subsubsection{Visualization of Local-aware Agents}\label{sec:Visualization of Mined Agents}
To further explore and understand the role of the local-aware agents, Fig.\ref{fig:10} exhibits the visualization of activation maps of different agents on the query image, which are derived by performing similarity computation between the agents and the query features. We could observe the following phenomena: (i) Different agents are able to focus on different yet complementary local regions, which strongly demonstrates the effectiveness of the local semantic mining as conceived. (ii) The constructed agents can reasonably focus on key semantic clues in the current episode, e.g., in the first column, different agents will focus on the `Fuselage', `Wings', and `Tail' of the multiple `Planes'. And this rationalization is more obvious in simpler natural scenarios (PASCAL-$5^i$). This also proves the generalizability of the proposed method from another perspective.

\begin{figure*}[t]
\setlength{\abovecaptionskip}{1pt} \setlength{\belowcaptionskip}{1pt}
\centering
\includegraphics[width=1.0\linewidth]{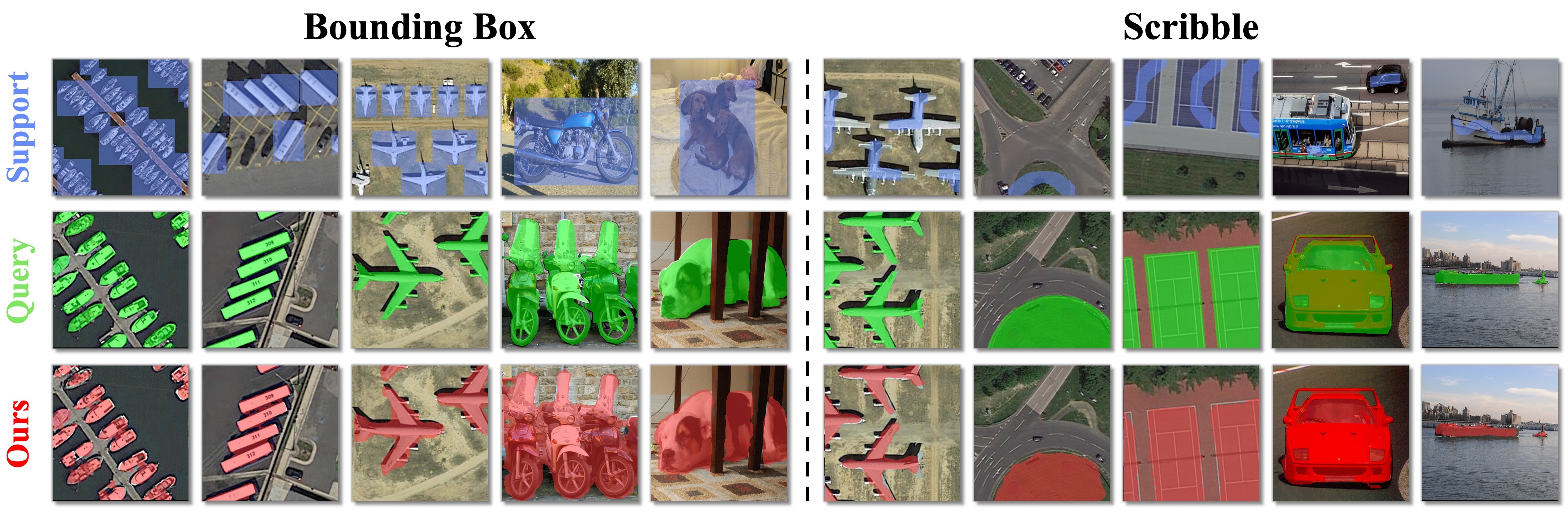}
\caption{\textbf{Qualitative visualization results for Weak-label Few-shot Segmentation under the 1-shot setting.} The left panel shows the results of employing bounding box masks, while the right panel shows the results of employing scribble masks. From top to bottom, each row represents: support images with GT (\textcolor{blue}{blue}), query images with GT (\textcolor{green}{green}), our segmentation results (\textcolor{red}{red}).}
\label{fig:15}
\end{figure*}

\begin{table*}[t]
\setlength{\abovecaptionskip}{1pt}
\setlength{\belowcaptionskip}{1pt}
\caption{Extended experiments of Weak-label Few-shot Segmentation on three FSS benchmarks measured in mIoU (\%) under the 1-shot setting. `Dense' denotes the densely labeled mask. `Bounding box' and `Scribble' denote the bounding box label and scribble label, respectively.} \label{tab:6}
\renewcommand\arraystretch{1.65}
\centering
\resizebox{1.0\linewidth}{!}{
\begin{tabular}{r|c|ccc|c|cccc|c|cccc|c} 
\toprule
\multirow{2}{*}{Method}                & \multirow{2}{*}{Scheme} & \multicolumn{4}{c|}{iSAID}       & \multicolumn{5}{c|}{PASCAL-$5^i$}         & \multicolumn{5}{c}{COCO-$20^i$}            \\ 
\cline{3-16}
&                         & Fold-0 & Fold-1 & Fold-2 & Mean  & Fold-0 & Fold-1 & Fold-2 & Fold-3 & Mean  & Fold-0 & Fold-1 & Fold-2 & Fold-3 & Mean   \\ 
\hline
DCPNet (IJCV2023)                      & Dense                   & 48.43  & 37.59  & 52.09  & 46.04 & 67.20  & 72.90  & 65.20  & 59.40  & 66.10 & 43.00  & 48.60  & 45.40  & 44.80  & 45.50  \\
FPTrans (NeurIPS2022)                  & Dense                   & 51.59  & 37.79  & 52.85  & 47.41 & 72.30  & 70.60  & 68.30  & 64.10  & 68.83 & 44.40  & 48.90  & 50.60  & 44.00  & 46.98  \\ 
\hline
\multirow{3}{*}{\textbf{AgMTR (Ours)}} & \textbf{Bbox}           & 55.19  & 38.98  & 48.70  & 47.62 & 59.03  & 64.51  & 55.44  & 56.49  & 58.81 & 32.65  & 43.40  & 38.14  & 43.50  & 39.42  \\
& \textbf{Scribble}       & 56.20  & 40.75  & 49.68  & 48.88 & 67.92  & 73.45  & 62.30  & 64.69  & 67.09 & 37.22  & 49.84  & 45.73  & 44.97  & 44.43  \\ 
\cline{2-16}
& Dense                   & 57.51  & 43.54  & 53.69  & 51.58 & 72.13  & 75.72  & 66.31  & 66.91  & 70.27 & 44.01  & 55.51  & 49.90  & 46.61  & 49.01  \\
\bottomrule
\end{tabular}}
\end{table*}

\subsection{Extension in Weak-label Few-shot Segmentation}\label{sec:Extension in Weak-label Few-shot Segmentation}
To further alleviate the burden of dense labeling in the real world, this section discusses the effectiveness of FSS with two cheaper mask schemes, bounding box and scribble. Following the setup of~\citep{wang2019panet}, the above cheaper schemes are randomly generated from the original dense labeled masks of support images, as shown in Fig.\ref{fig:15}. Table \ref{tab:6} demonstrates the performance of different mask schemes on three benchmarks under the 1-shot setting. One can find that our method with the scribble mask can achieve 48.88\%, 67.09\%, and 44.43\% mIoU accuracy on three benchmarks, respectively, which performs comparably with several competing methods with the dense mask, revealing AgMTR's tolerance to label quality. In addition, we notice that employing the bounding box mask provides weaker results than the scribble mask, which may be due to the fact that the bounding box mask contains a large amount of background noise that interferes with the mining of valuable semantics from the support images by AgMTR. Based on the above observations, we believe that the scribble mask can effectively alleviate the labeling burden while maintaining excellent performance.

\begin{figure*}[t]
\setlength{\abovecaptionskip}{1pt} \setlength{\belowcaptionskip}{1pt}
\centering
\includegraphics[width=1.0\linewidth]{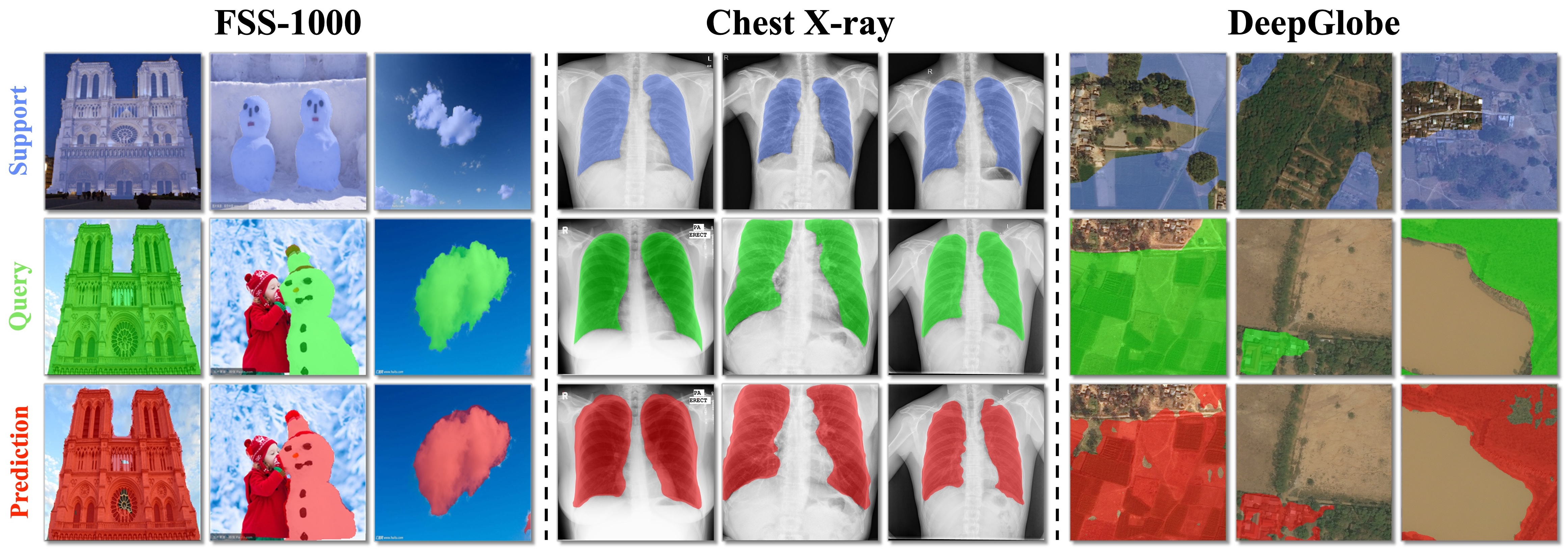}
\caption{\textbf{Qualitative visualization results for Cross-domain Few-shot Segmentation under the 1-shot setting,} where the model is trained on PASCAL VOC dataset and evaluated on three datasets, including FSS-1000, Chest X-ray, and DeepGlobe.}
\label{fig:15_5}
\end{figure*}

\begin{table*}[t]
\setlength{\abovecaptionskip}{1pt}
\setlength{\belowcaptionskip}{1pt}
\caption{Extended experiments of Cross-domain FSS on three datasets. Notably, the proposed model is trained on PASCAL VOC and evaluated on these three datasets. Some comparison results are from \citep{herzog2024adapt} and \citep{lei2022cross}.} \label{tab:6_6}
\renewcommand\arraystretch{1.4}
\centering
\resizebox{1.0\linewidth}{!}{
\begin{tabular}{r|cccc|cccc} 
\toprule
\multirow{2}{*}{Method} & \multicolumn{4}{c|}{1-shot}                                       & \multicolumn{4}{c}{5-shot}                                        \\ 
\cline{2-9}
& FSS-1000        & X-ray    & Deepglobe      & Average        & FSS-1000        & X-ray    & Deepglobe      & Average        \\ 
\hline
PFENet (TPAMI2021)~\citep{tian2020prior}      & 70.87          & 27.22          & 16.88          & 38.32          & 70.52          & 27.57          & 18.01          & 38.70          \\
RePRI (CVPR2021)~\citep{boudiaf2021few}        & 70.96          & 65.08          & 25.03          & 53.69          & 74.23          & 65.48          & 27.41          & 55.71          \\
HSNet (ICCV2021)~\citep{min2021hypercorrelation}        & 77.53          & 51.88          & 29.65          & 53.02          & 80.99          & 54.36          & 35.08          & 56.81          \\
PATNet (ECCV2022)~\citep{lei2022cross}       & 78.59          & 66.61          & 37.89          & 61.03          & 81.23          & 70.20          & 42.97          & 64.80          \\
HDMNet (CVPR2023)~\citep{peng2023hierarchical}       & 75.10          & 30.60          & 25.40          & 43.70          & 78.60          & 31.30          & 39.10          & 49.67          \\
RestNet (BMVC2023)~\citep{huang2023restnet}      & 81.50          & 70.40          & 22.70          & 58.20          & \underline{84.90}  & 73.70          & 29.90          & 62.83          \\
PMNet
(WACV2024)~\citep{chen2024pixel}      & \textbf{84.60} & 70.40          & 37.10          & 64.03          & \textbf{86.30} & 74.00          & 41.60          & 67.30          \\
ABCDFSS (CVPR2024)~\citep{herzog2024adapt}      & 74.60          & \underline{79.80}  & \textbf{42.60} & \underline{65.67}  & 76.20          & \underline{81.40}  & \textbf{49.00} & \underline{68.87}  \\ 
\hline
\rowcolor{gray!15} \textbf{AgMTR (Ours)}   & \underline{83.54}  & \textbf{86.50} & \underline{41.42}  & \textbf{70.49} & 84.36          & \textbf{88.85} & \underline{47.11}  & \textbf{73.44}          \\
\bottomrule
\end{tabular}}
\end{table*}

\subsection{Extension in Cross-domain Few-shot Segmentation}\label{sec:Extension in Cross-domain Few-shot Segmentation}

To explore the generalization of the proposed AgMTR to unseen domains in real-world situations, we perform the Cross-domain Few-shot Segmentation, where the samples utilized for training and testing are from different domains. Following the setup of ABCDFSS~\citep{herzog2024adapt} and PATNet~\citep{lei2022cross}, we evaluate the performance on three datasets with different domain shifts, including FSS-1000~\citep{li2020fss}, Chest X-ray~\citep{jaeger2014two}, and DeepGlobe~\citep{demir2018deepglobe}, which cover 1,000 classes of everyday objects, X-ray images and 6 classes of satellite images, respectively. The proposed method is trained on PASCAL VOC and evaluated on these three datasets, with the mIoU results and visualization results presented in Table \ref{tab:6_6} and Fig.\ref{fig:15_5}. It can be found that our method achieves satisfactory performance, e.g., AgMTR exceeds ABCDFSS by 8.94\% on FSS-1000, and by 6.7\% on the Chest X-ray under the 1-shot setting. The above results demonstrate that even though our AgMTR is not specifically constructed for cross-domain scenarios, the designed local agents can dynamically mine class-specific semantics from labeled, unlabeled, and query images in the target domain to perform accurate segmentation and exhibit excellent generalisability.

\begin{figure}[t]
\setlength{\abovecaptionskip}{1pt} \setlength{\belowcaptionskip}{1pt}
\centering
\includegraphics[width=1.0\linewidth]{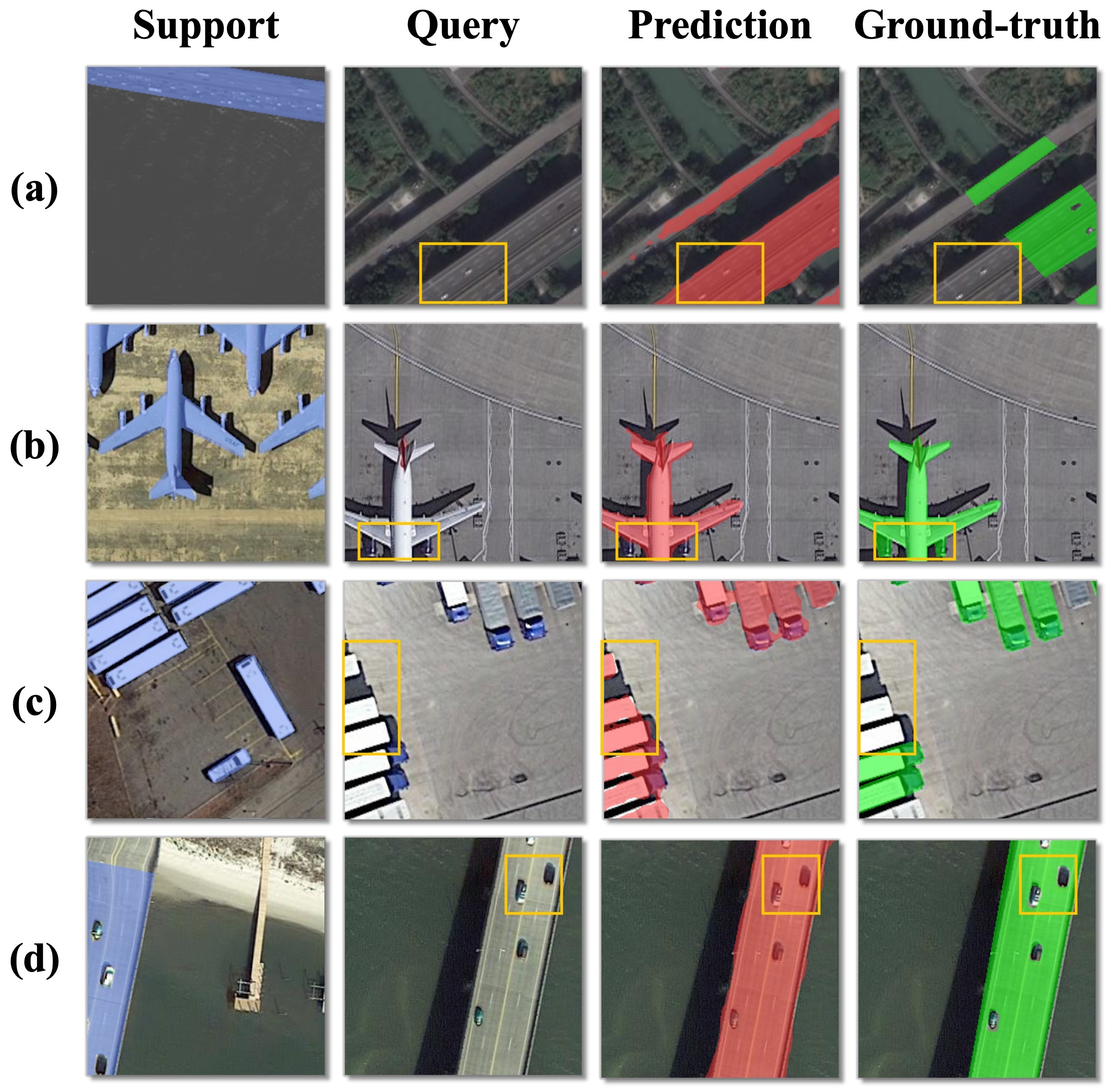}
\caption{\textbf{Failure cases of the proposed method.} From top to bottom: (a) the `Roadway' on the land which is misinterpreted. (b) The `Engine
in the shadow. (c) The similarity between `Large vehicles' and `Containers'. (d) Overlap between `Vehicles' and the `Bridge'.}
\label{fig:14}
\end{figure}

\subsection{Failure case analysis}\label{sec: Failure case analysis}
By mining and optimizing local-aware agents from support, unlabeled, and query images, AgMTR is able to construct precise agent-level semantic correlation to efficiently perceive the interested object. However, it ignores the association between foreground objects and background, or other categories, producing several failure segmentation cases. Take the case (a) in Fig.\ref{fig:14} as an example, the model fails to understand the `Bridge is always over water' connection and incorrectly considers a `Road' over the land as a `Bridge'. In addition, when the object appears in the shadow, the object shows extreme similarity with the background, or the object overlaps with other classes, the segmentation performance is also weak (see Fig.\ref{fig:14}.(b)-(d)). A potential solution to radically improve the FSS performance may be to utilize the diffusion models to generate more labeled images based on the support image to provide more robust guidance, such as DifFSS~\citep{tan2023diffss}. 

\section{Conclusion}
To correct the semantic ambiguity between the query FG and BG pixels due to pixel-level mismatch, this paper proposes a novel Agent Mining Transformer (AgMTR) for remote sensing FSS, which can adaptively mine a set of representative agents to construct agent-level semantic correlation. Compared to pixel-level semantics, the agents are equipped with local contextual information and possess a broader receptive field, thus ensuring that the query pixels safely execute the semantic aggregation and enhancing the semantic clarity between the query FG and BG pixels. Experiment results on remote sensing scenarios iSAID and more common natural scenarios, i.e., PASCAL-$5^i$ and COCO-$20^{i}$ validate the effectiveness of the proposed AgMTR and set the state-of-the-art performance, which strongly demonstrates its generality and extensibility. In the future, we consider introducing the semantic information of linguistic modality in FSS tasks, such as Visual-Language models, and try to explore more challenging Zero-shot segmentation tasks.

\section*{Data Availability Statement}
The datasets generated in this study are available from the iSAID\footnote{\url{https://captain-whu.github.io/iSAID/index}} website, the PASCAL\footnote{\url{http://host.robots.ox.ac.uk/pascal/VOC/}} website, and the MS COCO\footnote{\url{https://cocodataset.org/\#home}} website.

\section*{Acknowledgments}
This work was supported by the National Natural Science Foundation of China (NSFC) under Grant 62301538.

\bibliographystyle{spbasic}      
\bibliography{3dfsl_main}   

\clearpage

\end{sloppypar}  
\end{document}